\documentclass[10pt,twocolumn,letterpaper]{article}

\usepackage[pagenumbers]{cvpr} 

\usepackage{booktabs}
\usepackage{enumitem}
\usepackage{array}
\usepackage[dvipsnames]{xcolor}

\usepackage{graphicx}
\usepackage{amsmath}
\usepackage{amssymb}
\usepackage{booktabs}
\usepackage{multirow}
\usepackage[accsupp]{axessibility}
\usepackage{colortbl}  
\usepackage{xcolor}

\definecolor{cvprblue}{rgb}{0.21,0.49,0.74}
\usepackage[pagebackref,breaklinks,colorlinks,citecolor=cvprblue]{hyperref}

\newcommand{\mj}{$\mathcal{J}$}
\newcommand{\mf}{$\mathcal{F}$}
\newcommand{\mjf}{$\mathcal{J}\&\mathcal{F}$}

\newcommand{\mjs}{$\mathcal{J}_s$}
\newcommand{\mfs}{$\mathcal{F}_s$}
\newcommand{\mju}{$\mathcal{J}_u$}
\newcommand{\mfu}{$\mathcal{F}_u$}
\newcommand{\mg}{$\mathcal{G}$}
\newcommand{\std}[1]{\tiny{$\pm$#1}}

\def\methodNAME{GLEE\xspace}
\title{General Object Foundation Model for Images and Videos at Scale}

\author{Junfeng Wu$^{1}$\textsuperscript{\textnormal{*}},
~~Yi Jiang$^{2}$\textsuperscript{\textnormal{*}},~~Qihao Liu$^{3}$, ~~Zehuan Yuan$^{2}$,~~Xiang Bai$^{1}$\textsuperscript{\textnormal{$\dagger$}},~~Song Bai$^{2}$\textsuperscript{\textnormal{$\dagger$}}\\
{\fontsize{10.5pt}{12pt}\selectfont $^{1}$Huazhong University of Science and Technology, $^{2}$ByteDance Inc., $^{3}$Johns Hopkins University}\\  
}

\begin{document}
\maketitle

\begin{abstract} 
\def\thefootnote{*}\footnotetext{Equal Technical Contribution. $^\dagger$Correspondence to Xiang Bai $<$\url{xbai@hust.edu.cn}$>$ and Song Bai $<$\url{songbai.site@gmail.com}$>$.}

We present \methodNAME in this work, an object-level foundation model for locating and identifying objects in images and videos. Through a unified framework, \methodNAME accomplishes detection, segmentation, tracking, grounding, and identification of arbitrary objects in the open world scenario for various object perception tasks.
Adopting a cohesive learning strategy,
\methodNAME acquires knowledge from diverse data sources with varying supervision levels to formulate general object representations, excelling in zero-shot transfer to new data and tasks.
Specifically, we employ an image encoder, text encoder, and visual prompter to handle multi-modal inputs, enabling  to simultaneously solve various object-centric downstream tasks while maintaining state-of-the-art performance. Demonstrated through extensive training on over five million images from diverse benchmarks, \methodNAME exhibits remarkable versatility and improved generalization performance, efficiently tackling downstream tasks without the need for task-specific adaptation.
By integrating large volumes of automatically labeled data, we further enhance its zero-shot generalization capabilities. 
Additionally, \methodNAME is capable of being integrated into Large Language Models, serving as a foundational model to provide universal object-level information for multi-modal tasks. We hope that the versatility and universality of our method will mark a significant step in the development of efficient visual foundation models for AGI systems. The model and code will be released at \url{https://glee-vision.github.io/}.

\end{abstract}    
\section{Introduction}
\label{sec:intro}

\begin{figure}[tb]
\centering
\includegraphics[width = 0.45\textwidth]{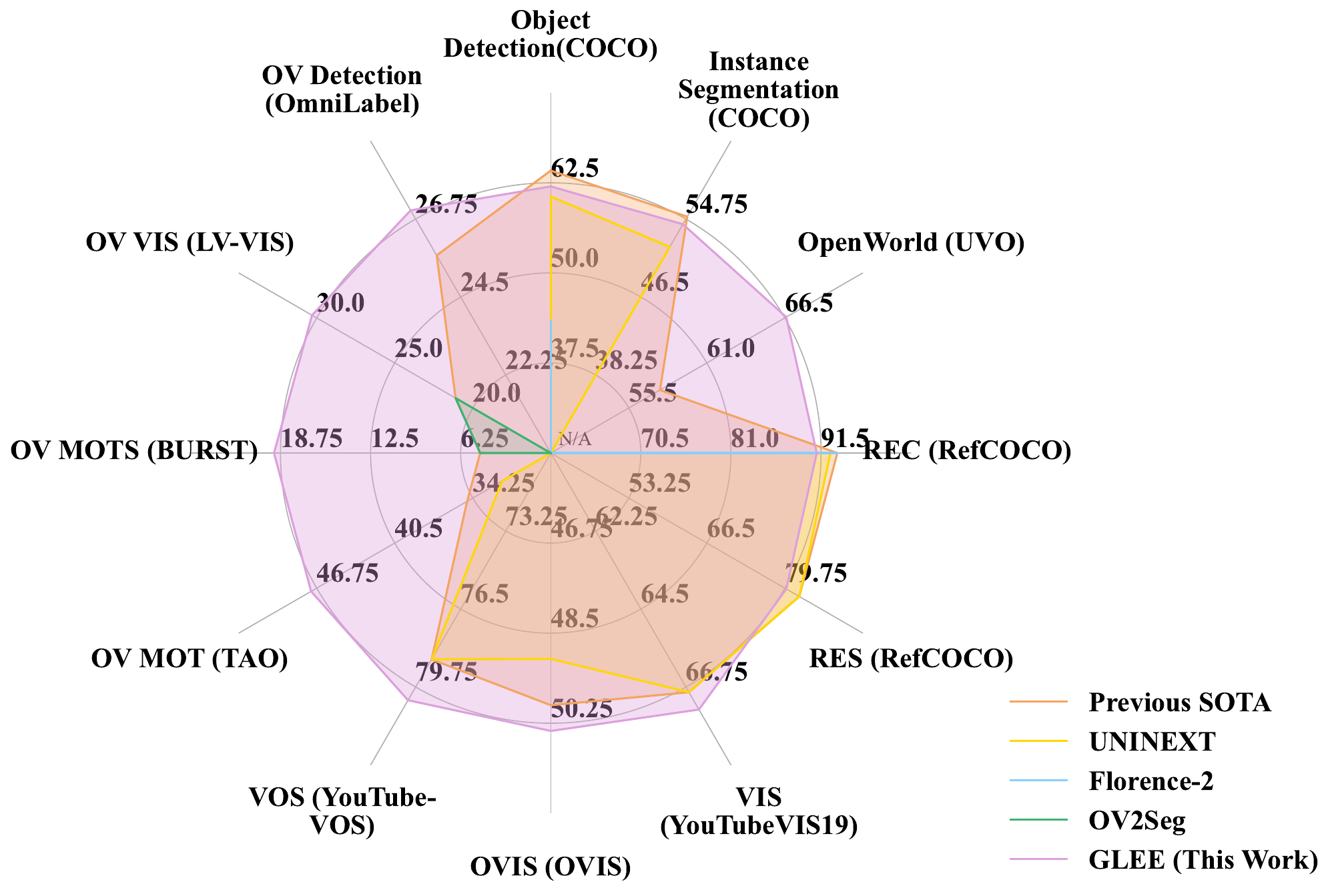}
\caption{
The performance of \methodNAME on a broad range of object-level tasks compared with existing models.
}
\vspace{-2ex}
\label{fig:intro}
\end{figure}

Foundation model~\cite{bommasani2021opportunities} is an emerging paradigm for building artificial general intelligence (AGI) systems,
signifying a model trained on broad data that is capable of being adapted to a wide range of downstream tasks in an general paradigm.
Recently, NLP foundation models such as BERT~\cite{bert}, GPT-3~\cite{gpt3}, T5~\cite{T5} developed with unified input-output paradigms and large-scale pre-training, have achieved remarkable generalization capabilities to address nearly all NLP tasks.

In computer vision, the diversity of task types and the lack of a unified from makes visual foundation models only serve specific subdomains, such as CLIP~\cite{CLIP} for multi-modal visual model, MAE~\cite{MAE} for visual representations model, SAM~\cite{SAM} for segmentation model.
Despite being widely studied, current visual foundation models are still focusing on establishing correlations between global image features and language descriptions or learning image-level feature representations.
However, locating and identifying objects constitute foundational capabilities in computer vision systems, serves as a basis for solving complex or high level vision tasks such as segmentation, scene understanding, object tracking, event detection, and activity recognition and support a wide range of applications.

In this work, we advance the development of object-level foundation models within the visual domain.
To address the aforementioned limitation, providing general and accurate object-level information, we introduce a general object visual foundation model, coined as \methodNAME, which simultaneously solve a wide range of object-centric tasks while ensuring SOTA performance, including object detection, instance segmentation, grounding, object tracking, interactive segmentation and tracking, etc., as shown in Figure~\ref{fig:intro}.
Through a unified input and output paradigm definition, our model is capable of learning from a wide range of diverse data and predicting general object representations, which masks it to generalize well to new data and tasks in a zero-shot manner and achieve amazing performance.
In addition, thanks to the unified paradigm, the training data can be scaled up at low cost by introducing a large amount of automatically labeled data, and further improve the zero-shot generalization ability of the model.

\textbf{A general object foundation model framework. }
Our objective is to build an object visual foundation model capable of simultaneously addressing a wide range of object-centric tasks. Specifically, we employ an image encoder, a text encoder, and a visual prompter to encode multi-modal inputs. 
They are integrated into a detector to extract objects from images according to textual and visual input. This unified approach to handle multiple modalities enables us to concurrently solve various object-centric tasks, including detection~\cite{coco, detr, sparsercnn, deformableDETR}, instance segmentation~\cite{MaskRCNN, mask2former}, referring expression comprehension~\cite{RIS, seqtr, polyformer, uniref}, interactive segmentation~\cite{polygon-rnn, polygon-rnn++, SEEM}, multi-object tracking~\cite{motchallenge, CenterTrack, Trackformer, bytetrack, unicorn}, video object segmentation~\cite{youtubevos, stm, stcn, xmem}, video instance segmentation~\cite{MaskTrackRCNN, vistr, seqformer, IDOL, GenVIS}, and video referring segmentation~\cite{referformer, uniref, urvos}, all while maintaining state-of-the-art performance.

\begin{figure*}[tb]
\centering
\includegraphics[width = 0.99\textwidth]{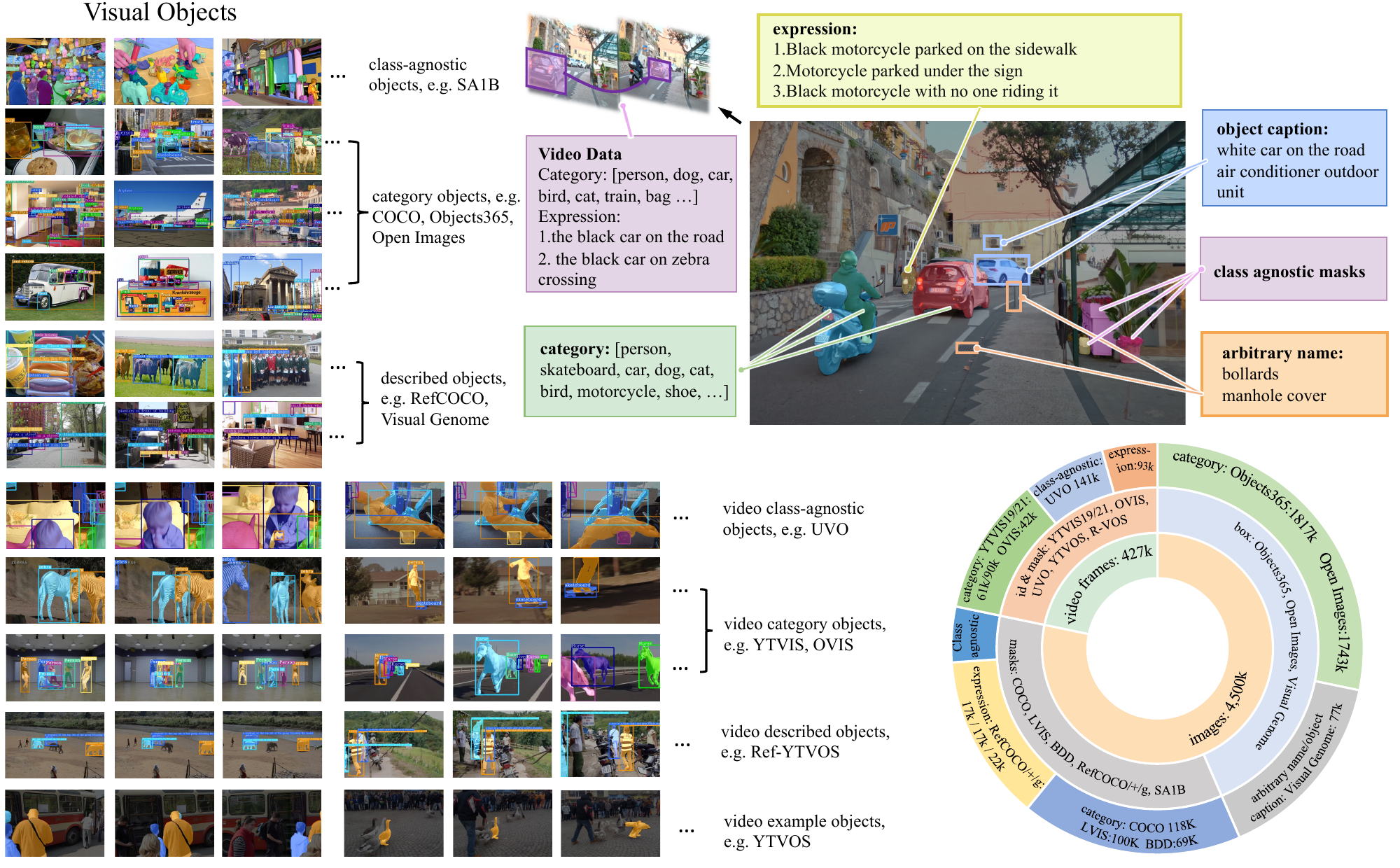}
\vspace{-2ex}
\caption{
An illustrative example showcasing annotations of varying granularities from different datasets, along with the scale of data we utilized. Training on datasets from multiple sources endows the model with more universal representations.
}
\label{fig:datademo}
\end{figure*}

\textbf{A multi-granularity joint supervision and scaleable training paradigm. 
}
The design of the unified framework capable of addressing multiple tasks enables joint training on over five million images from diverse benchmarks and varying levels of supervision. Existing datasets differ in annotation granularity: detection datasets like Objects365~\cite{objects365} and OpenImages~\cite{OpenImages} offer bounding boxes and category names; COCO~\cite{coco} and LVIS~\cite{lvis} provide finer-grained mask annotations; RefCOCO~\cite{RefCOCOandplus,RefCOCOg-umd} and Visual Genome ~\cite{visualgenome} provide detailed object descriptions. Additionally, video data enhance the temporal consistency of model, while open-world data contribute class-agnostic object annotations. 
An intuitive display of the supervision types and data scales of the datasets employed is presented in  Figure~\ref{fig:datademo}.
The unified support for multi-source data in our approach greatly facilitates the incorporation of additional manually or automatically annotated data, enabling easy scaling up of the dataset.
Furthermore, the alignment of model optimization across tasks means that joint training serves not only as a unifying strategy but also as a mechanism to boost performance across individual tasks.


\textbf{Strong zero-shot transferability to a wide range of object level image and video tasks.} 
After joint training on data from diverse sources, \methodNAME demonstrates remarkable versatility and zero-shot generalization abilities. Extensive experiments demonstrate that \methodNAME achieves state-of-the-art performance compared to existing specialist and generalist models in object-level image tasks such as detection, referring expression comprehension, and open-world detection, all without requiring any task-specific designs or fine-tuning.
Furthermore, we showcase the extraordinary generalization and zero-shot capabilities of \methodNAME in large-vocabulary open-world video tracking tasks, achieving significantly superior performance over existing models even in a zero-shot transfer manner.
Additionally, by incorporating automatically annotated data like SA1B~\cite{SAM} and GRIT~\cite{kosmos2}, we are able to scale up our training dataset to an impressive size of 10 million images at a low cost, which is typically challenging to achieve for object-level tasks and further enhances the generalization performance.
Moreover, we replace the SAM~\cite{SAM} component with \methodNAME in a multimodal Large Language Model (mLLM)~\cite{lai2023lisa} and observe that it achieves comparable results. 
This demonstrates that \methodNAME is capable of supplying the visual object-level information that modern LLMs currently lack, thus laying a solid foundation for an object-centric mLLMs.

\section{Related Work}
\label{sec:Related}

\subsection{Visual Foundation Model}

As foundation models~\cite{bert, gpt3, T5, PALM, llama} in the NLP field have achieved remarkable success, the construction of visual foundation models attracts increasing attention. Unlike NLP tasks that are predominantly unified under a text-to-text paradigm, tasks in Computer Vision still exhibit significant differences in form and definition. This disparity leads to visual foundation models typically being trained in a single-task learning frameworks, limiting their applicability to tasks within certain sub-domains. For instance, multi-modal visual foundation models like CLIP~\cite{CLIP}, ALIGN~\cite{ALIGN}, Florence~\cite{florence}, BEIT3~\cite{beit3}, Flamingo\cite{flamingo} make significant advancements in efficient transfer learning and demonstrate impressive zero-shot capabilities on vision-language tasks by employing contrastive learning and masked data modeling on large-scale image-text pairs. DALL-E~\cite{DALLE,DALLE2} and Stable Diffusion~\cite{stable_diffusion} are trained on massive pairs of images and captions, enabling them to generate detailed image content conditioned on textual instruction. DINO~\cite{sslDINO}, MAE~\cite{MAE}, EVA~\cite{fang2023eva}, ImageGPT~\cite{imageGPT} obtain strong visual representations through self-supervised training on large-scale image data, which are then employed to transfer to downstream tasks.
These foundation models learn image-level features, which are not directly applicable to object-level tasks. 
The recently proposed SAM~\cite{SAM}, capable of segmenting any object of a given image based on visual prompt such as points and boxes, provides rich object-level information and demonstrates strong generalization capabilities. However, the object information lacks semantic context, limiting its application in object-level tasks. Unlike existing visual foundation models, we aim to develop an object foundation model that directly solve downstream tasks without the need for additional parameters or fine-tuning.

\subsection{Unified and General Model}
Unified models share similarities with foundation models in the aspect of multi-task unification for their ability to handle multiple vision or multi-modal tasks within a single model. 
MuST~\cite{MuST} and Intern~\cite{Intern} propose to train across multiple vision tasks and solving them simultaneously. 
Inspired by the success of sequence-to-sequence NLP models~\cite{gpt3,T5}, models such as Uni-Perceiver~\cite{Uni-perceiver}, OFA~\cite{OFA}, Unified-IO~\cite{unified-io}, Pix2Seq v2~\cite{Pix2Seqv2}, and UniTAB~\cite{unitab} propose modeling various tasks as sequence generation tasks within a unified paradigm.
While these approaches have demonstrated promising cross-task generalization capabilities, they focus mainly on image-level understanding tasks. 
In addition, their auto-regressive generation of boxes and masks results in significantly slower inference speeds and the performance still falls short of state-of-the-art task-specific models.
Building upon on detectors~\cite{maskdino,deformableDETR}, Uni-Perceiver v2~\cite{Uni-perceiverv2} and UNINEXT~\cite{UNINEXT} utilize unified maximum likelihood estimation and object retrieval to support various tasks, effectively resolves the challenges of localization. Nonetheless, they are trained on closed-set data, thereby not exhibiting zero-shot generalization capabilities. X-decoder~\cite{xdecoder} and  SEEM~\cite{SEEM} construct a generalized decoding model capable of predicting pixel-level segmentation and language tokens. 
Diverging from unified models, the proposed \methodNAME not only directly addresses object-level tasks in a unified manner but also provides universal object representations, which generalize well to new data and tasks, serving as a cornerstone for a broader range of tasks that require detailed object information.

\begin{figure*}[tb]
\centering
\includegraphics[width=0.99 \linewidth]{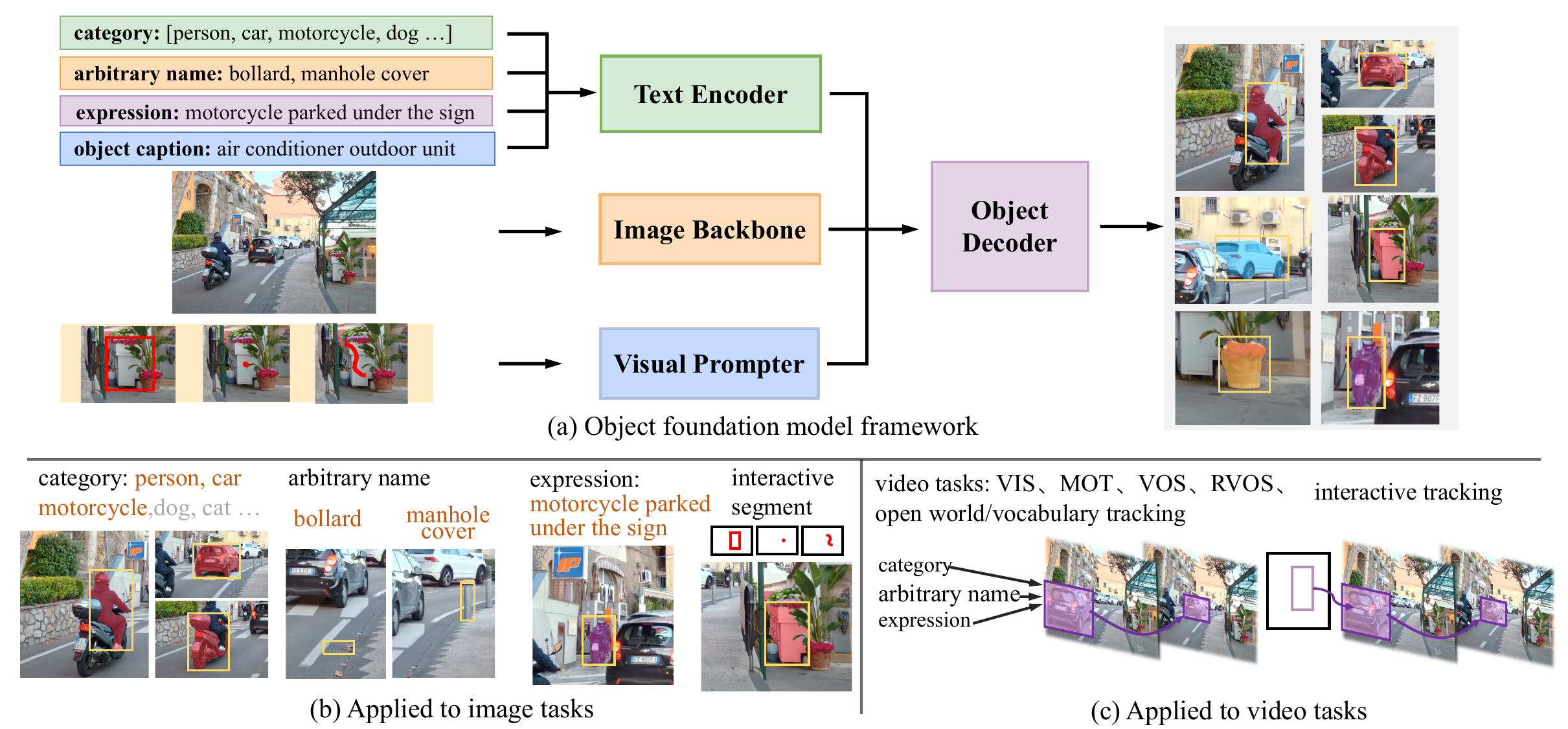}
\caption{
\textbf{Framework of \methodNAME. }
The text encoder accepts textual descriptions in various forms from diverse data sources, including object categories, names, captions, and referring expressions.
The visual prompter encodes points, bounding boxes, or scribbles into corresponding visual representations.The object decoder take them and image features to predict objects in images.
(b) illustrates the application of \methodNAME to image tasks tailored for different language descriptions and visual prompts. (c) demonstrates the application across various object-level video tasks.
}
\label{fig:pipeline}
\vspace{-2ex}
\end{figure*}

\subsection{Vision-Language Understanding}
Open-vocabulary detection (OVD) and Grounding models both necessitate the localization and recognition of as many objects as possible. With the recent advancements in vision-language pre-training \cite{CLIP, ALIGN, florence, yu2022coca}, a commonly employed strategy for OVD involves transferring the knowledge from pre-trained vision-language models (VLMs) to object detectors~\cite{gu2021open, minderer2022simple, kuo2023open}.
Another group of studies leverages extensive image-text pair datasets to broaden the detection vocabulary \cite{zareian2021open, zhong2022regionclip, feng2022promptdet, GLIP, lin2022learning, yao2023detclipv2}. 
However, these language-based detectors are inherently constrained by the capabilities and biases of language models, making it challenging to excel simultaneously in both localization and recognition. Our objective is to optimally utilize existing datasets to construct a general object-level foundation model, aims to not only detect and identify objects effectively but also to offer universal object representations for a wide range of downstream tasks


\section{Method}
\label{sec:Method}

 \subsection{Formulation}

The proposed \methodNAME consists of an image encoder, a text encoder, a visual prompter, and an object decoder, as illustrated in Figure~\ref{fig:pipeline}.
The text encoder processes arbitrary descriptions related to the task, including object categories, names in any form, captions about objects, and referring expressions. The visual prompter encodes user inputs such as points, bounding boxes, or scribbles during interactive segmentation into corresponding visual representations of target objects. 
Then they are integrated into a detector for extracting objects from images according to textual and visual input.
We build the object decoder upon MaskDINO~\cite{maskdino} with a dynamic class head by compute similarity between object embedding from detector and text embedding from the text encoder.
Given an input image $I \in \mathcal{R}^{3\times H\times W}$, we first extract multi-scale features $Z$ with backbones such as ResNet~\cite{resnet}. Then we feed them into the object decoder
and adopt three prediction heads (classification, detection, and segmentation) on the output embedding $q_d \in \mathcal{R}^{N\times C}$ from decoder. 
Following other object segmentation models~\cite{ViTDet,mask2former, maskdino}, we construct a 1/4 resolution pixel embedding map $M_p \in \mathcal{R}^{C \times \frac{H}{4} \times \frac{W}{4}} $ which is obtained by upsampling and fusing multi-scale feature maps from the backbone and Transformer encoder.
Finally, we obtain each binary mask prediction $m \in \mathcal{R}^{N \times \frac{H}{4} \times \frac{W}{4}} $ via a dot product between the
N mask embeddings and pixel embedding map:
\begin{align}
    \vspace{-2ex}
    m = FFN(q_d) \otimes M_p,
    \label{eq:1}
    \vspace{-2ex}
\end{align}
where FFN is a 3-layer feed forward head with ReLU activation function and a linear projection layer.

To support arbitrary vocabularies and object descriptions, we replace the FFN classifier with text embeddings following DetCLIP~\cite{detclip}.
Specifically,
we feed K category names as separate sentences into the text encoder $Enc_L$ and use the average of each sentence tokens as the output text embedding $e_t \in \mathcal{R}^{K\times D}$ for each category or description. 
Then we compute the alignment scores
$S_{align} \in \mathcal{R}^{N\times K}$ between object embedding and text embedding:
\begin{align}
    S_{align} = q_d \cdot W_{i2t} \otimes e_t,
    \label{eq:2}
\end{align}
where $W_{i2t} \in \mathcal{R}^{C\times D} $ is image-to-text projection weights.
We use logits $S_{align}$ to replace traditional classification logits to compute Hungarian matching cost during training and assign categories to objects during inference.
To make the original visual features prompt-aware, an early fusion module is adopted before Transformer encoder following UNINEXT~\cite{UNINEXT}, which tasks image feature from backbone and prompt embedding as input and perform bi-directional cross-attention between them.

\subsection{Task Unification}

Based on the above designs, \methodNAME can be used to seamlessly unify
a wide range of object perception tasks in images and videos,
including object detection, instance segmentation, grounding, multi-target tracking (MOT), video instance segmentation (VIS), video object segmentation (VOS), interactive segmentation and tracking, and supports open-world/large-vocabulary image and video detection and segmentation tasks.

\textbf{Detection and Instance Segmentation.}
For detection task, a fixed-length category list is given and all objects in the category list are required to be detected. 
For a dataset with category list length K,
the text input can be formulated as $\{p_k\}^{K}_{k=1}$ where $p_k$ represents for the k-th category name, e.g., P = [“person”, “bicycle”, “car”, ... , “toothbrush”] for COCO~\cite{coco}.
For datasets with large vocabulary, calculating the text embedding of all categories is very time-consuming and redundant. Therefore, for datasets with a category number greater than 100, such as objects365~\cite{objects365} and LVIS~\cite{lvis}, suppose there are $\hat{K}$ positive categories in an image, we take the  $\hat{K}$ positive categories and then pad the category number to 100 by randomly sampling from the negative categories. For instance segmentation, we enable the mask branch and add mask matching cost with mask loss.

\textbf{Grounding and Referring Segmentation.}
These tasks provide reference textual expressions, where objects are described with attributes, for example,Referring Expression Comprehension (REC)~\cite{RefCOCOandplus,seqtr}, Referring Expression Segmentation (RES)~\cite{RefCOCOandplus,polyformer}, and Referring Video Object Segmentation (R-VOS)~\cite{urvos, referformer} 
aim at finding objects matched with the given language expressions like “The fourth person from the left”. 
For each image, we take the all the object expressions as text prompt and feed the them into the text encoder. For each expressions, we apply global average pooling along the sequence dimension to get text embedding $e_t$. 
The text embeddings are feed into early fusion module and additionally interacte with object queries through self-attention module in the decoder. 

\textbf{MOT and VIS.}
Both Multi-object Tracking (MOT)\cite{motchallenge, Tracktor, CenterTrack, Trackformer, bytetrack} and Video Instance Segmentation (VIS)\cite{MaskTrackRCNN, ovis, IDOL, GenVIS} need to detect and track all the objects in the predefined category list, and VIS requires additional mask for the objects. These two tasks can be considered as extended tasks of detection and instance segmentation on videos. 
We found that with sufficient image exposure, object embeddings from the decoder effectively distinguish objects in a video, showing strong discriminability and temporal consistency.
As a result, they can be directly employed for tracking without the need for an additional tracking head.
Training on image-level data can address straightforward tracking scenarios, but in cases of severe occlusion scenes, such as OVIS~\cite{ovis}, image-level training cannot guarantee that the model exhibits strong temporal consistency under occlusion conditions. Therefore, for occlusion scenarios, it is essential to utilize video data for training. Following IDOL~\cite{IDOL}, we sample two frames from a video and introduce contrastive learning between frames to make the embedding of the same object instance closer in the embedding space, and the embedding of different object instances farther away.
During Inference, the detected objects are tracked by simple bipartite matching of the corresponding object queries following MinVIS~\cite{MinVIS}.

\textbf{Visual Prompted Segmentation.}
Interactive segmentation~\cite{boykov2001interactive, rother2004grabcut, wu2014milcut, sofiiuk2022reviving, xu2016deep, polygon-rnn, simpleclick} takes various forms of visual prompt, such as points, boxes, or scribbles, to segment the specified objects within an image. On the other hand, VOS aims to segment the entire object throughout the entire video based on a mask provided in the first frame of the video.
We extract visual prompt embeddings twice in the model.
First, we crop the prompt square area from RGB image and send it to the backbone to obtain the visual prompt feature of the corresponding area, and send it to the early fusion module before the Transformer encoder.
Second, we sample fine-grained visual embeddings from the pixel embedding map $M_p$ according to visual prompt and make them interacted with object queries through self-attention module in the Transformer decoder layer, as the same with text embeddings.

\begin{table*}[!ht]
\centering
\resizebox{1.0\linewidth}{!}{
\begin{tabular}{lcccccccccccccccc} 
\toprule
\multirow{3}{*}{Method}   & \multirow{3}{*}{Type}     & \multicolumn{6}{c}{ {\it{Generic Detection \& Segmentation}}}     & \multicolumn{6}{c}{ {\it{Referring Detection \& Segmentation}}}       & \multicolumn{1}{c}{ {\it{OpenWorld}}}         \\
 \cmidrule(lr){3-10} \cmidrule(lr){11-16} \cmidrule(lr){17-17}

&    & \multicolumn{2}{c}{COCO-val}  & \multicolumn{2}{c}{COCO-test-dev}    & \multicolumn{4}{c}{LVIS}   &  \multicolumn{2}{c}{RefCOCO}    & \multicolumn{2}{c}{RefCOCO+}      & \multicolumn{2}{c}{RefCOCOg}  & \multicolumn{1}{c}{UVO}  \\
 \cmidrule(lr){3-4}   \cmidrule(lr){5-6}   \cmidrule(lr){7-10}  \cmidrule(lr){11-12}  \cmidrule(lr){13-14}  \cmidrule(lr){15-16}    \cmidrule(lr){17-17} 
&  &$\rm AP_{box}$  & $\rm AP_{mask}$ &$\rm AP_{box}$  & $\rm AP_{mask}$  & $\rm AP_{box}$ & $\rm AP_{r-box}$  &$\rm AP_{mask}$  &$\rm AP_{r-mask} $ & $\rm P@0.5 $ &$\rm  oIoU$  & $\rm P@0.5 $ & $\rm oIoU $ & $\rm P@0.5 $ & $\rm oIoU $      &$\rm AR_{mask} $       \\ 
\hline
MDETR~\cite{mdetr}       & \multirow{8}{*}{Specialist}    & -   & -   & - & -   & -   & -   & -   & -     &87.5   & -   & 81.1   & -   & 83.4    & -    & -\\
SeqTR~\cite{seqtr}       &\multirow{8}{*}{Models}       & -   & -   & - & -   & -   & -   & -   & -    &87.0  &71.7  &78.7  &63.0  &82.7   & 64.7   & -    \\
PolyFormer (L)~\cite{polyformer}  &   &-   & - & -   & -   & -   & -   & -   & -     &90.4   & 76.9   & 85.0   & 72.2   & 85.8    & 71.2 & -  \\
ViTDet-L ~\cite{ViTDet}   &   &57.6  &49.8 &-   & -   &51.2  &-  &46.0  &34.3  & -   & -   & -   & -   & -   & -  & -\\
ViTDet-H ~\cite{ViTDet}  &   &58.7  &50.9 &-   & -   &53.4  &-  &48.1   &36.9  & -   & -   & -   & -   & -   & - & - \\
EVA-02-L~\cite{eva02} &    &64.2  &55.0   &64.5  &55.8 &65.2  &-   &57.3  &-   \\
ODISE~\cite{ODISE}    &     & -   & - & -   & -   & -   & -   & -   & -     & -   & -   & -   & -   & -   & -   &57.7  \\
Mask2Former (L)~\cite{mask2former}   &     & -   & 50.1   & -   & 50.5   & -   & -   & -   & -   & -     & -   & -   & -   & -   & -   & - \\
MaskDINO (L)~\cite{maskdino}   &     & -   & 54.5   & -   & 54.7    & -   & -   & -   & -   & -     & -   & -   & -   & -   & -   & -  \\
\midrule
UniTAB (B)~\cite{unitab}           & \multirow{12}{*}{Generalist}   &-  & -  & -   & -    & -   & -   & -   & -     &88.6   & -   & 81.0   & -   & 84.6    & -  & - \\ 
OFA (L)~\cite{OFA}  & \multirow{12}{*}{Models}  &-  & -   & -   & -   & -   & -   & -   & -     &90.1   & -   & 85.8   & -   & 85.9    & - & - \\
Pix2Seq v2 ~\cite{Pix2Seqv2}    &     &46.5   &38.2   & -   & -    & -   & -   & -   & -    & -   & -   &-   & -   & -   & -    & -  \\ 
Uni-Perceiver-v2 (B)~\cite{Uni-perceiverv2}  &     &58.6   &50.6   & -   & - & -   & -   & -   & -     & -  & -   & -   & -   &-     & - & -   \\
Uni-Perceiver-v2 (L)~\cite{Uni-perceiverv2} &     &61.9  & 53.6  & -   & -    & -   & -   & -   & -     & -  & -   & -   & -   &-     & - & -  \\
UNINEXT (R50)~\cite{UNINEXT} &   &51.3   &44.9    & -   & -   &36.4   & -   & -   & -     & 89.7  & 77.9   & 79.8   & 66.2   &84.0    & 70.0  & -  \\
UNINEXT (L)~\cite{UNINEXT} &     &58.1   &49.6   & -   & -   & -   & -   & -   & -     & 91.4   & 80.3   & 83.1   &70.0    & 86.9   & 73.4 & -\\
UNINEXT (H)~\cite{UNINEXT} &     &60.6   &51.8   & -   & -   & -   & -   & -   & -     & 92.6   & 82.2   & 85.2   &72.5    & 88.7  & 74.7 & - \\
GLIPv2 (B)~\cite{GLIPv2} &   &-   & - & 58.8   & 45.8  & -   & -   & -   & -   & -     & -   & -   & -   & -   & -   & - \\
GLIPv2 (H)~\cite{GLIPv2} &   &-   & - & 60.6   & 48.9  & -   & -   & -   & -   & -     & -   & -   & -   & -   & -   & - \\
X-Decoder (B)~\cite{xdecoder}  & &- &45.8 &- &45.8  & -   & -   & -   & -   & -   & -  & -   & -   & -  & -   & -   \\
X-Decoder (L)~\cite{xdecoder}  & &- &46.7 &- &47.1   & -   & -   & -   & -   & -   & -  & -   & -   & -  & -   & -   \\
Florence-2 (L)~\cite{florence2}  &     &43.4   & -   & -  & -   & -   & -   & -   & -     &93.4   & -   & 88.3   & -   & 91.2    & - & - \\
\midrule
\methodNAME-Lite  & \multirow{2}{*}{Foundation}    
                        &55.0   &48.4  &54.7  &48.3  &44.2  &36.7  &40.2  &33.7    &88.5   &77.4   &78.3    &64.8   &82.9   &68.8    &66.6 \\
\methodNAME-Plus  & \multirow{2}{*}{Models}     &60.4   &53.0  &60.6 &53.3  &52.7  &44.5  &47.4  &40.4     &90.6   &79.5   &81.6    &68.3   &85.0   &70.6   &70.6 \\
\methodNAME-Pro   &     &62.0   &54.2  &62.3 &54.5   &55.7  &49.2  &49.9  &44.3       &91.0   &80.0   &82.6    &69.6   &86.4   &72.9  &72.6  \\

\bottomrule
\end{tabular}
}
\caption{Comparison of \methodNAME to recent specialist and generalist models on object-level image tasks.
For REC and RES tasks, we report Precision@0.5 and overall IoU (oIoU).
For open-world instance segmentation task, we reported the average recall of 100 mask proposals (AR@100) on the UVO~\cite{UVO}.
}
\label{tab:all_image_tasks}
\vspace{-2ex}
\end{table*}

\subsection{Training Unification}
\label{TrainingUnification} 
\textbf{Tasks with Dynamic Loss.}
We jointly train \methodNAME in an end-to-end manner on over 5 million images from diverse benchmarks with various levels of supervision. Different loss functions are selected for training on various datasets.
There are six types of losses for our \methodNAME: semantic loss, box loss, mask loss, confidence loss, contrastive tracking loss, and distillation loss.
For all tasks with category list or object expressions, we apply focal loss~\cite{focalloss} as semantic loss on logits $S_{align}$ to align the text concepts with object features.
For box prediction, we use a combination of L1 loss and generalized IoU loss~\cite{giou}. The mask loss is defined as a combination of the Dice loss~\cite{diceloss} and Focal loss. 
For the Visual Prompt Segmentation tasks, we employ an additional FFN to predict the confidence score for each object queries supervised by focal loss.
Following IDOL~\cite{IDOL}, for video tasks, we sample two frames and apply contrastive tracking loss on the object query from the last layer of decoder:

\begin{align}
    \mathcal{L}_{embed} = \log [1+\sum_{\textbf{k}^+}\sum_{\textbf{k}^-}\exp(\textbf{v} \cdot \textbf{k}^-  - \textbf{v} \cdot \textbf{k}^+) ],
    \label{eq:4}
\vspace{-2ex}
\end{align}
where $\textbf{k}^+$ \text{and} $\textbf{k}^-$ are the object queries belong to the same object and other objects from the reference frame, respectively.
For the text encoder, we distill the knowledge from the teacher CLIP text encoder
to ensure the text embedding in pre-trained vison-language embedding space.
We apply an L1 loss between our text encoder and CLIP text encoder to minimize their distance:
\begin{align}
    \mathcal{L}_{text} = \frac{1}{K} \sum_{i=0}^K \left\lVert Enc_{L}(p_i) - Enc_{CLIP}(p_i)   \right\rVert.
    \label{eq:4}
    \vspace{-2ex}
\end{align}

\begin{table*}[!ht]
\centering
\resizebox{1.0\linewidth}{!}{
\begin{tabular}{lccccccccccccccc} 
\toprule
\multirow{3}{*}{Method}      & \multicolumn{4}{c}{ {\it{Tracking Any Object (TAO~\cite{tao})}}}     & \multicolumn{6}{c}{ {\it{BURST~\cite{burst}}}}       & \multicolumn{3}{c}{ {\it{LV-VIS~\cite{LVVIS}}}}         \\
 \cmidrule(lr){2-5} \cmidrule(lr){6-11} \cmidrule(lr){12-14} 

 &  \multirow{2}{*}{$\rm TETA$}   &\multirow{2}{*}{$\rm LocA$}   &\multirow{2}{*}{$\rm AssocA$}  &\multirow{2}{*}{$\rm ClsA$}     &  \multicolumn{2}{c}{ALL}    & \multicolumn{2}{c}{Common}      & \multicolumn{2}{c}{Uncommon}  &\multirow{2}{*}{$\rm AP$}  &\multirow{2}{*}{$\rm AP_b$}  &\multirow{2}{*}{$\rm AP_n$}   \\
  \cmidrule(lr){6-7}  \cmidrule(lr){8-9}  \cmidrule(lr){10-11}   
&  & &  & &$\rm HOTA$  &$\rm mAP$  &$\rm HOTA$  &$\rm mAP$     &$\rm HOTA$  &$\rm mAP$       \\ 
\hline
Tracktor~\cite{bergmann2019tracking}                    & 24.2  & 47.4 & 13.0  & 12.1 &-&-&-&- &-&-&-&- &-\\
DeepSORT~\cite{wojke2017simple}                         & 26.0  & 48.4 & 17.5  & 12.1 &-&-&-&- &-&-&-&- &-\\
Tracktor++~\cite{tao}                        & 28.0  & 49.0 & 22.8  & 12.1 &-&-&-&- &-&-&-&- &-\\
QDTrack~\cite{QDTrack}                       & 30.0  & 50.5 & 27.4  & 12.1 &-&-&-&- &-&-&-&- &-\\
TETer~\cite{li2022tracking}                            & 33.3  & {51.6} & 35.0  & 13.2 &-&-&-&- &-&-&-&- &-\\ 
OVTrack$\dagger$  ~\cite{li2023ovtrack}        & 34.7  & 49.3 & {36.7}  & 18.1 &-&-&-&- &-&-&-&- &-\\
STCN Tracker$\dagger$~\cite{burst}   &-&-&-&-  & 5.5 & 0.9 &17.5  & 0.7   &2.5  &0.6  &-&- &-\\
Box Tracker$\dagger$~\cite{burst}    &-&-&-&-  & 8.2  &1.4  &27.0  &3.0   &3.6   &0.9 &-&- &-\\
Detic~\cite{detic}-SORT$\dagger$~\cite{bewley2016simple}   &-&-&-&- &-&-&-&- &-&- & 12.8 & 21.1 &6.6      \\
Detic~\cite{detic}-XMem~$\dagger$\cite{xmem}  &-&-&-&- &-&-&-&- &-&- & 16.3  &24.1 &10.6  \\
OV2Seg-R50$\dagger$~\cite{LVVIS}  &-&-&-&- &-&3.7 &-&- &-&-  &14.2 &17.2 &11.9\\
OV2Seg-B$\dagger$~\cite{LVVIS}     &-&-&-&- &-&4.9&-&- &-&-  &21.1 &27.5 &16.3 \\
UNINEXT (R50)~\cite{UNINEXT} &31.9  &43.3  &35.5   &17.1  &-&-&-&- &-&-&-&- &-\\    
\midrule
\methodNAME-Lite$\dagger$  &40.1  &56.3  &39.9  &24.1 &22.6  &12.6  &36.4  &18.9   &19.1   &11.0    &19.6 &22.1 &17.7 \\
\methodNAME-Plus$\dagger$  &41.5  &52.9  &40.9  &\textbf{30.8}    &26.9  &17.2  &38.8  &23.7  &23.9   &15.5   &\textbf{30.3} &\textbf{31.6} &\textbf{29.3}  \\
\methodNAME-Pro$\dagger$  &\textbf{47.2}  &\textbf{66.2}  &\textbf{46.2}  &29.1   &\textbf{31.2}   &\textbf{19.2}   &\textbf{48.7}   &\textbf{24.8 }  &\textbf{26.9}  &\textbf{17.7}  &23.9  &24.6   &23.3\\
 
\bottomrule
\end{tabular}
}
\caption{Comparison of \methodNAME to recent specialist and generalist models on object-level video tasks in a zero-shot manner. Evaluation metrics of BURST are reported separately for ‘common’, ‘uncommon’ and ‘all’ classes. 
The mAP computes mask IoU at the track level, HOTA is a balance of per-frame detection accuracy (DetA) and temporal association accuracy (AssA), and TETA that deconstructs detection into localization and classification components.
The $\rm AP$, $\rm AP_b$, and $\rm AP_n$ in LV-VIS mean the average precision of overall categories, base categories, and novel categories. $\dagger$ does not use videos for training. The under-performance of Pro relative to Plus on LV-VIS is due to Pro employing larger training and inference resolutions, which prove to be sub-optimal for this specific dataset}
\label{tab:three_video_zeroshot}
\end{table*}

\textbf{Data Scale Up.}
A visual foundation model should be able to easily scale up the training data and achieve better generalization performance. 
Thanks to the unified training paradigm, the training data can be scaled up at low cost by introducing a large amount of automatically labeled data from SA1B~\cite{SAM} and GRIT~\cite{kosmos2}.
SA1B provides large and detailed mask annotations, which enhance the object perception capabilities of model, while GRIT offers a more extensive collection of referring-expression-bounding-box pairs, improving the object identification abilities and the understanding capability of descriptions. Ultimately, we introduced 2 million SA1B data points and 5 million GRIT data points into the training process, bringing the total training data to 10 million.

\section{Experiments}

\subsection{Experimental Setup}
\textbf{Datasets and Training Strategy.} We conducted training in three stages.
Initially, we performed pretraining for object detection task on Objects365~\cite{objects365} and OpenImages~\cite{OpenImages}, initializing the text encoder with pretrained CLIP~\cite{CLIP} weights and keeping the parameters frozen. 
In the second training step, we introduced additional instance segmentation datasets, including COCO~\cite{coco}, LVIS~\cite{lvis}, and BDD~\cite{bdd100k}. Furthermore, we treat three VIS datasets: YTVIS19~\cite{MaskTrackRCNN}, YTVIS21~\cite{ytvis21dataset}, and OVIS~\cite{ovis}, as independent image data to enrich the scenes. For datasets that provide descriptions of objects, we included RefCOCO~\cite{RefCOCOandplus}, RefCOCO+~\cite{RefCOCOandplus}, RefCOCOg~\cite{RefCOCOg-umd}, Visual Genome~\cite{visualgenome}, and RVOS~\cite{urvos}. Since Visual Genome contains multiple objects in a single image, we treat it as detection task and used both object descriptions and object noun phrases as categories, with a total of 200 dynamic category lists per batch. Additionally, we introduced two open-world instance segmentation datasets, UVO~\cite{UVO} and a subset of SA1B~\cite{SAM}. For these two datasets, we set the category name for each object to be 'object' and train in instance segmentation paradigm.
During the second step, text encoder is unfrozen but supervised by distillation loss to ensure the predicted text embedding in CLIP embedding space.
After the second step, \methodNAME demonstrated state-of-the-art performance on a range of downstream image and video tasks and exhibited strong zero-shot generalization capabilities, unless otherwise specified, all the experimental results presented below were obtained by the model at this stage. 

Building upon this, we introduce the SA1B and GRIT datasets to scale up the training set, resulting in a model named \textbf{\methodNAME-scale}, which exhibited even stronger zero-shot performance on various downstream tasks. 
Since image data alone is insufficient for the model to learn temporal consistency features, we incorporated sequential video data from YTVIS, OVIS, RVOS, UVO, and VOS to improve its performance if specifically note.

\textbf{Implementation Details.}
In our experiments, 
we developed \methodNAME-Lite, \methodNAME-Plus, and \methodNAME-Pro using ResNet-50~\cite{resnet}, Swin-Large~\cite{SwinTransformer}, and EVA-02 Large~\cite{eva02} as the vision encoder respectively.
Following MaskDINO~\cite{maskdino}, we adopt deformable transformer in object decoder, and use 300 object queries. Query denoising and Hybrid matching are kept to accelerate convergence and improve performance.
During pretraining, we set a minibatch to 128 on 64 A100 GPUs, for 500,000 iterations.
For joint-training, we train \methodNAME on 64 A100 GPUs for 500,000 iterations, further training details, data pre-processing methods, and data sampling strategies can be found in the supplementary materials.
More detailed information on data usage and model training is available in the supplementary materials.

\begin{table*}[ht]

\begin{center}
\resizebox{\linewidth}{!}{
\begin{tabular}{l@{\hskip9pt}| 
l@{\hskip9pt}l@{\hskip9pt}l@{\hskip9pt} 
l@{\hskip9pt}l@{\hskip9pt}l@{\hskip9pt}
l@{\hskip9pt}l@{\hskip9pt}l@{\hskip9pt}l@{\hskip9pt}
l@{\hskip9pt}l@{\hskip9pt}l@{\hskip9pt}l@{\hskip9pt}l@{\hskip9pt}l@{\hskip9pt}l}
\toprule

Model  & \small{PascalVOC} &
\small{AerialDrone} & 
\small{Aquarium} &
\small{Rabbits} &
\small{EgoHands} &
\small{Mushrooms} &
\small{Packages} &
\small{Raccoon} &
\small{Shellfish} &
\small{Vehicles} &
\small{Pistols} &
\small{Pothole} &
\small{Thermal} & 
Avg
\\
\midrule

 GLIP-T 
 & 56.2
& 12.5
& 18.4
& 70.2
& 50.0
& 73.8
& 72.3
& 57.8
& 26.3
& 56.0
& 49.6
& 17.7
& 44.1
  & 46.5
  
\\

 GLIP-L  
 & 61.7
& 7.1
& 26.9
& 75.0
& 45.5
& 49.0
& 62.8
& 63.3
& 68.9
& 57.3
& 68.6
& 25.7
& 66.0
& 52.1
\\
 
\methodNAME-Lite
& 61.7
& 7.9 
&23.2 
& 72.6 
& 41.9 
& 51.6 
& 32.9 
& 51.1 
& 35.0 
& 59.4 
& 45.6 
& 21.8 
& 56.9 
& 43.2
\\

\methodNAME-Lite-Scale
& 61.2 
& 5.0 
& 23.9 
& 71.9 
& 46.2 
& 57.8 
& 25.6 
& 56.8 
& 33.1 
& 60.6 
& 57.1 
& 25.3 
& 52.5 
& 44.4 
\\

\methodNAME-Plus
& 67.8 
& 10.8 
& 38.3 
& 76.1 
& 47.4 
& 19.2 
& 29.4 
& 63.8 
& 66.7 
& 63.8 
& 62.6 
& 15.3 
& 66.5 
& 48.3 
\\

\methodNAME-Plus-Scale
& 67.5 
& 12.1 
& 39.7 
& 75.8 
& 50.3 
& 41.1 
& 42.4 
& 66.4 
& 64.0 
& 62.8 
& 61.8 
& 17.5 
& 63.8 
& 51.2 
\\

\methodNAME-Pro
& 68.9 
& 16.5 
& 37.6 
& 77.2 
& 23.3 
& 40.1 
& 44.7 
& 68.2 
& 66.2 
& 66.1 
& 63.2 
& 18.1 
& 65.8 
& 50.5 
\\

\methodNAME-Pro-Scale
& 69.1 
& 13.7 
& 34.7 
& 75.6 
& 38.9 
& 57.8 
& 50.6 
& 65.6 
& 62.7 
& 67.3 
& 69.0 
& 30.7 
& 59.1 
& 53.4
\\

\bottomrule
\end{tabular}
}
\vspace{-1ex}
\caption{Zero-shot performance on 13 ODinW datasets.}
\label{table:odinw_zero_shot}
\vspace{-2ex}
\end{center}
\end{table*}

\begin{table}[!ht]
\centering
\resizebox{1.0\linewidth}{!}{
\setlength{\tabcolsep}{2pt}
\begin{tabular}{llcccccc}
\toprule 
\multirow{2}{*}{Method}  &\multirow{2}{*}{Backbone} & \multicolumn{3}{c}{YTVIS 2019 val~\cite{MaskTrackRCNN}}  & \multicolumn{3}{c}{OVIS val~\cite{ovis}} \\
\cmidrule(lr){3-5} \cmidrule(lr){6-8}
&&$\rm AP$    &$\rm AP_{50}$  &$\rm AP_{75}$ &$\rm AP$    &$\rm AP_{50}$  &$\rm AP_{75}$\\

\midrule
IFC~\cite{IFC}   & \multirow{9}{*}{ResNet-50}   &42.8 &65.8 &46.8&13.1&27.8&11.6 \\  
SeqFormer~\cite{seqformer}   &     &47.4 &69.8 &51.8 &15.1 &31.9 &13.8\\
IDOL~\cite{IDOL}  &   & 49.5 &{74.0} &52.9&{30.2}&{51.3}&{30.0}\\
VITA~\cite{VITA}& &{49.8}&72.6&{54.5}&19.6&41.2&17.4\\
GenVIS~\cite{GenVIS} &  &51.3  &72.0  &57.8  &34.5  &59.4  &35.0    \\
DVIS~\cite{DVIS}  &  &52.6  &76.5  &58.2  &34.1  &59.8  &32.3    \\
NOVIS~\cite{NOVIS}   &  &52.8 &75.7 &56.9 &32.7 &56.2 &32.6    \\
UNINEXT  &   &{53.0}& {75.2}& {59.1} & {34.0} & {55.5} & {35.6}\\  
\textbf{\methodNAME-Lite} &  &53.1  &74.0  &59.3  &27.1/32.3  &45.4/52.2  &26.3/33.7    \\ 
\midrule
SeqFormer~\cite{seqformer}  &  \multirow{7}{*}{Swin-L}  &{59.3}  &{82.1}  &{66.4} & - & - & -\\ 
VITA~\cite{VITA}& &63.0&86.9&67.9&27.7&51.9&24.9\\
IDOL~\cite{IDOL}  &  &64.3 &{87.5} &71.0&{42.6} &65.7&{45.2}\\  
GenVIS~\cite{GenVIS} &  &63.8  &85.7  &68.5  &45.4  &69.2  &47.8   \\
DVIS~\cite{DVIS}  &  &64.9 &\textbf{88.0} &72.7 &49.9 &\textbf{75.9} &53.0     \\
NOVIS~\cite{NOVIS}   &  &65.7 &87.8 &72.2   &43.5 &68.3 &43.8       \\
\textbf{\methodNAME-Plus} & &63.6 &85.2 &70.5    &29.6/40.3  &50.3/63.8  &28.9/39.8 \\ 
\midrule
UNINEXT  &ConvNeXt-L &64.3 &87.2 &71.7 &41.1 &65.8 &42.0 \\
UNINEXT  &ViT-H  &66.9 &87.5 &\textbf{75.1} &49.0 &72.5 &52.2  \\
\textbf{\methodNAME-Pro} &EVA02-L  &\textbf{67.4}  &87.1  &74.1  &38.7/\textbf{50.4}  &59.4/71.4  &39.7/\textbf{55.5}    \\ 
\bottomrule
\end{tabular}}

\caption{
Performance comparison of our \methodNAME on video instance segmentation tasks.
}
\vspace{-2ex}
\label{table:vis_results}
\end{table}

\subsection{Comparison with Generalist Models}
We demonstrate the universality and effectiveness of our model as an object-level visual foundation model, directly applicable to various object-centric tasks while ensuring state-of-the-art performance without needing fine-tuning. 
We compare our approach with existing specialist and generalist models in image-level tasks, including detection, referring expression comprehension, and open-world instance segmentation.
We report detection and instance segmentation results on both the COCO validation~\cite{coco} set and LVIS val v1.0~\cite{lvis}. 
While sharing almost identical image sets, LVIS is distinguished by its annotations of over 1,200 object categories, showcasing a long-tail distribution. This distinction makes LVIS more representative of challenging real-world scenarios due to its broader category coverage. As indicated in Table~\ref{tab:all_image_tasks}, our model outperforms all generalist models on both COCO and LVIS benchmarks. Even when compared to other state-of-the-art specialist approaches, which are tailored with specific design, our model remains highly competitive.
This demonstrates that \methodNAME, while mastering universal and general object representations, concurrently maintains advanced capabilities in object detection and segmentation. This characteristic is vitally important for adapting to a broad spectrum of downstream tasks requiring precise object localization.
For the REC and RES tasks, we evaluated our model on RefCOCO~\cite{RefCOCOandplus}, RefCOCO+~\cite{RefCOCOandplus}, and RefCOCOg~\cite{RefCOCOg-umd}, as show in Table~\ref{tab:all_image_tasks}, \methodNAME achieved comparable results with SOTA specialist methods PolyFormer~\cite{polyformer}, demonstrating strong capability to comprehend textual descriptions and showcasing potential to adapt to a broader range of multi-modal downstream tasks. 
In open-world instance segmentation tasks, 
we treated "object" as the category name, instructing the model to identify all plausible instance in an image in a class-agnostic manner. 
\methodNAME outperforms previous arts ODISE~\cite{ODISE} by 8.9 points, demonstrating the capability of identifying all plausible instance that might be present in an open-world scenario.

\subsection{Zero-shot Evaluation Across Tasks}
\textbf{Zero-shot Transfer to Video Tasks.}
The proposed \methodNAME is capable of adapting to new data and even new tasks in a zero-shot manner, without the need for additional fine-tuning. We evaluate its zero-shot capability on three large-scale, large-vocabulary open-world video tracking datasets: TAO~\cite{tao}, BURST~\cite{burst}, and LV-VIS~\cite{LVVIS}. TAO comprises 2,907 high-resolution videos across 833 categories. BURST builds upon TAO, encompassing 425 base categories and 57 novel categories. LV-VIS offers 4,828 videos within 1,196 well-defined object categories.
These three benchmarks require the model to detect, classify, and track all objects in videos, while BURST and LV-VIS additionally require segmentation results from the model. 
In Table~\ref{tab:three_video_zeroshot}, we compare the performance of our proposed model with existing specialist models. \textbf{Notably, the \methodNAME here is from the second training stage, which has not been exposed to images from these three datasets nor trained on video-level data.} Despite these constraints, \methodNAME achieves state-of-the-art performance that significantly exceeds existing methodologies. Specifically, \methodNAME surpasses the previous best method OVTrack by 36.0\% in TAO, nearly triples the performance of the best baseline in BURST, and outperforms OV2Seg~\cite{LVVIS} by 43.6\% in LV-VIS. This outstanding performance strongly validates the exceptional generalization and zero-shot capabilities of \methodNAME in handling object-level tasks across a range of benchmarks and tasks.

We additionally provide performance comparison on classic video segmentation tasks, including VIS, VOS, and RVOS.
As shown in Table~\ref{table:vis_results}, on the YTVIS2019~\cite{MaskTrackRCNN} benchmark, our model achieves SOTA results across various model sizes, surpassing all specialist models with complex designs to enhance temporal capabilities and the video unified model UNINEXT~\cite{UNINEXT}. On the OVIS~\cite{ovis} benchmark, which features lengthy videos with extensive object occlusions where temporal capabilities of object features are particularly crucial, our model does not directly reach SOTA. However, after a few hours of simple fine-tuning, it still achieves SOTA performance. 
This further validates the versatility and generalization capabilities of our model. More details on zero-shot evaluations for video tasks and demonstrations of interactive segmentation and tracking can be found in the Sec~\ref{sec:supp_videotask} of supplementary materials.

\textbf{Zero-shot Transfer to Real-world Downstream Tasks.
}
To measure generalization
on diverse real-world tasks, we evaluate zero-shot performance on OmniLabel~\cite{omnilabel}, which is a benchmark for evaluating language-based object detectors and encourages complex and diverse free-form text descriptions of objects.
As show in Table~\ref{table:omni_results}, compared to language-based detectors trained on large-scale caption data, \methodNAME significantly outperforms previous models in P-categ. However, due to the limited captions in our training dataset, it scores lower in AP-descr. By incorporating a more diverse set of box-caption data from the GRIT~\cite{kosmos2} to sclae up our training set, the AP-descr can be elevated to a level comparable with existing models. 
We conduct additional experiments on the “Object Detection in the Wild” (ODinW) benchmark~\cite{ODinW}, which is a suite of datasets covering a wide range of domains. We report the average mAP on the subset of 13 ODinW detection datasets introduced in~\cite{GLIP}, and report the per-dataset performance in a zero-shot manner, as shown in Table~\ref{table:odinw_zero_shot}. 
\methodNAME performs better than GLIP~\cite{GLIP} on the average of 13 public datasets, showcasing its robust generalization capability.
Furthermore, it is evident that by introducing automatically labeled data at a low cost for scaling up the training data, the zero-shot capabilities can be further enhanced, this reveals that \methodNAME has greater potential through scale-up.
A more comprehensive report on the per-dataset few-shot performance on ODinW is available in the supplementary materials to assess the adaptability of \methodNAME to other datasets.

\begin{figure}[tb]
\centering
\includegraphics[width=0.95 \linewidth]{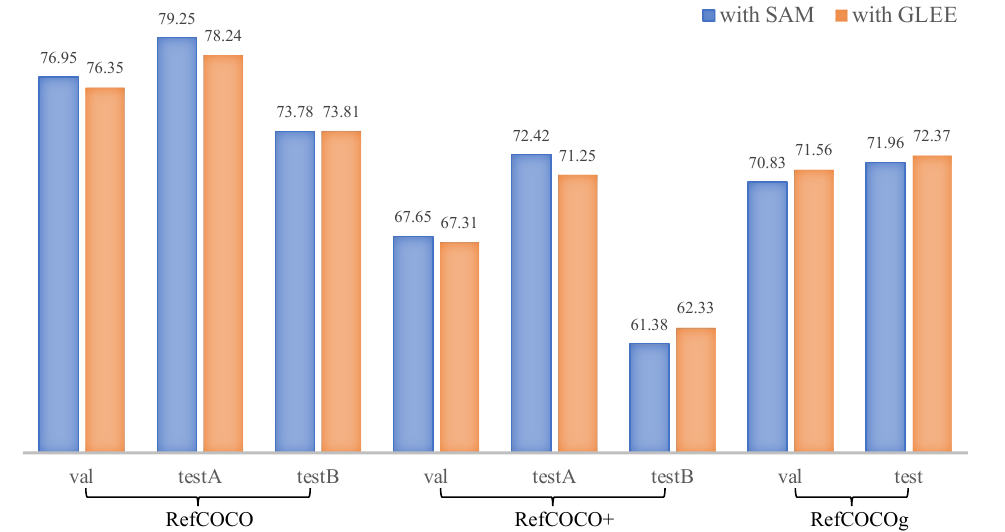}
\caption{
The performance comparison of replacing SAM with \methodNAME in LISA, \methodNAME achieves the same effectiveness as SAM in extracting objects.
}
\label{fig:lisa_sam}
\end{figure}

\begin{figure}[tb]
\centering
\includegraphics[width=0.95 \linewidth]{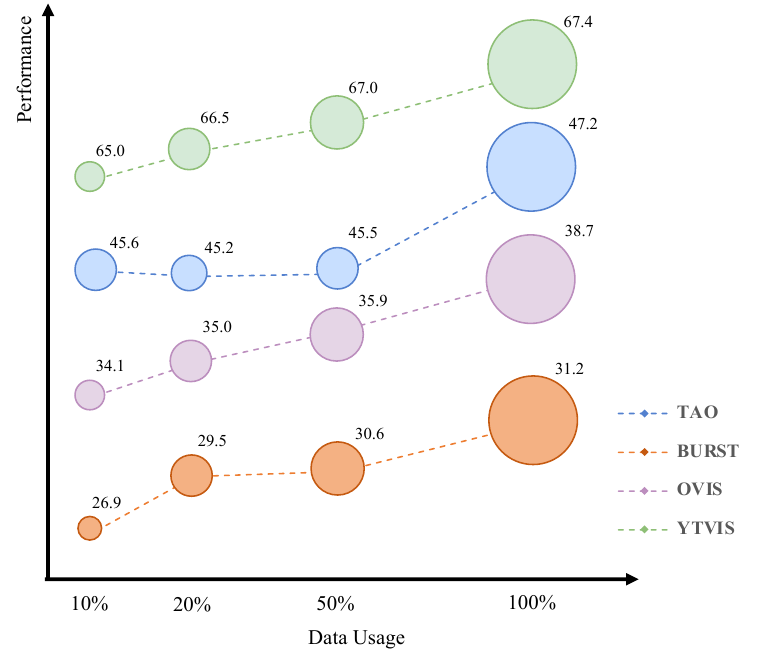}
\caption{
\textbf{Data scaling.} The performance of GLEE-Pro after training on 10\%, 20\%, 50\%, 100\% of the total data on TAO, BURST, OVIS, YTVIS19.
 Increased scale of training data result in enhanced zero-shot performance across diverse downstream tasks.
}
\label{fig:scaleup}
\end{figure}

\subsection{Serve as Foundation Model}
To explore whether \methodNAME can serve as a foundation model for other architectures, 
we selected LISA~\cite{lai2023lisa} for analysis, a mVLLM that combines LLAVA~\cite{llava} with SAM~\cite{SAM} for reasoning segmentation. We substituted its vision backbone with a frozen, pretrained \methodNAME-Plus and fed the object queries from \methodNAME into LLAVA and remove decoder of LISA. We directly dot product the output SEG tokens with \methodNAME feature map to generate masks. As shown in Figure~\ref{fig:lisa_sam}, after training for the same number of steps, our modified LISA-\methodNAME achieved comparable results to the original version, demonstrating the versatility of representations from \methodNAME and its effectiveness in serving other models.

\subsection{Ablation}

We conducted experiments to investigate the impact of training data scale on zero-shot performance across various tasks. To this end, we trained \methodNAME-Pro with 10\%, 20\%, 50\%, 100\% of the training data to evaluate the performance on zero-shot transfer tasks, including TAO, BURST, OVIS, and YTVIS as illustrated in the Figure~\ref{fig:scaleup}.
Our data scaling experiments reveal that increased sizes of training datasets result in enhanced zero-shot performance across diverse downstream tasks. This outcome implies that larger pre-training datasets are a valuable investment, offering a more effective and adaptable basis for a broad spectrum of downstream tasks. Thanks to the unified training approach of GLEE, we can efficiently incorporate any manually or automatically annotated data into our training process to achieve enhanced generalization capabilities.

\begingroup
\setlength{\tabcolsep}{1pt} 
\renewcommand{\arraystretch}{1.15} 
\begin{table}\centering
\footnotesize
\begin{tabular}{clccccccc}
\rotatebox{90}{Images} & \rotatebox{90}{Method} & \rotatebox{90}{AP} & \rotatebox{90}{AP-categ} & \rotatebox{90}{AP-descr} & \rotatebox{90}{AP-descr-pos} & \rotatebox{90}{AP-descr-S} & \rotatebox{90}{AP-descr-M} & \rotatebox{90}{AP-descr-L} \\
\toprule
\multirow{10}{*}{\rotatebox{90}{All}}
    & RegionCLIP~\cite{regionclip} & 2.7 & 2.7 & 2.6 & 3.2 & 3.6 & 2.7 & 2.3 \\  
    & Detic~\cite{detic} & 8.0 & 15.6 & 5.4 & 8.0 & 5.7 & 5.4 & 6.2 \\  
    \cline{2-9}
    & MDETR~\cite{mdetr} & - & - & 4.7 & 9.1 & 6.4 & 4.6 & 4.0 \\  
    & GLIP-T~\cite{GLIP} & 19.3 & 23.6 & 16.4 & 25.8 & 29.4 & 14.8 & 8.2 \\  
    & GLIP-L~\cite{GLIP} & 25.8 & 32.9 & 21.2 & 33.2 & 37.7 & 18.9 & 10.8 \\ 
    & FIBER-B~\cite{dou2022coarse} & 25.7 & 30.3 & 22.3 & 34.8 & 38.6 & 19.5 & 12.4 \\ 
    & \methodNAME-Lite & 20.3 & 37.5  & 14.0  & 19.1 & 23.0 & 12.7 & 10.0\\
    & \methodNAME-Lite-Scale & 22.7 & 35.5 & 16.7 & 22.3 & 33.7 & 14.3 & 10.2  \\
    & \methodNAME-Plus & 25.4 & 46.7 & 17.5 & 23.9 & 28.4 & 16.3 & 12.5    \\
    & \methodNAME-Plus-Scale & 27.0 & 44.5 & 19.4 & 25.9 & 36.0 & 17.2 & 12.4 \\
    \midrule    
\end{tabular}
\vspace{-1ex}
\caption{
Evaluation on the OmniLabel benchmark. The final AP value is the geometric mean of categories (AP-categ) and free-form descriptions (AP-descr).
}
    \vspace{-3ex}
\label{table:omni_results}
\end{table}
\endgroup


\section{Conclusion}
We introduce \methodNAME, a cutting-edge object-level foundation model designed to be directly applicable to a wide range of object-level image and video tasks. Crafted with a unified learning paradigm, \methodNAME learns from diverse data sources with varying levels of supervisions. \methodNAME achieves state-of-the-art performance on numerous object-level tasks and excels in zero-shot transfer to new data and tasks, showing its exceptional versatility and generalization abilities. Additionally, \methodNAME provides general visual object-level information, which is currently missing in modern LLMs, establishing a robust foundation for object-centric mLLMs.
{
    \small
    \bibliographystyle{ieeenat_fullname}
    \bibliography{main}
}

\clearpage
\setcounter{page}{1}
\maketitlesupplementary

In this supplementary material, we first provide more detailed information on data usage and model training in Sec~\ref{sec:supp_details}. Subsequently, in Sec~\ref{sec:supp_videotask}, we supplement additional zero-shot and  fine-tuning results on classic object-level video tasks, such as VOS and RVOS. In Sec~\ref{sec:supp_odinw}, detailed few-shot experimental results on the ODinW~\cite{ODinW} benchmark are provided to validate the transferability of \methodNAME to various real-world tasks. Finally, in Sec~\ref{sec:supp_interactive}, we showcase the results in interactive segmentation and tracking for images and videos.

\section{Datasets and Implementation Details}
\label{sec:supp_details}

\textbf{Data Preparation.}
To ensure the generalization of \methodNAME as an object-level foundation model, we conduct joint training using a substantial amount of data with region-level annotations from both images and videos. Existing datasets exhibit variations in annotation granularity: detection datasets such as Objects365~\cite{objects365} and OpenImages~\cite{OpenImages} provide bounding boxes and category names; COCO~\cite{coco} and LVIS~\cite{lvis} offer more detailed mask annotations; RefCOCO~\cite{RefCOCOandplus,RefCOCOg-umd} and Visual Genome~\cite{visualgenome} include comprehensive object descriptions. Furthermore, video datasets~\cite{MaskTrackRCNN,ytvis21dataset,ovis,urvos,youtubevos,UVO} contribute to the temporal consistency of models, and open-world data~\cite{SAM,UVO} enrich the annotations with class-agnostic object information. A comprehensive list of the datasets we utilized, along with their respective sizes and annotation granularities, is presented in Table~\ref{table:dataset_list}.
We extracted subsets of 500,000 and 2,000,000 images from the SA1B~\cite{SAM} dataset for joint training in stage 2 and scale-up training respectively.
To ensure that objects from SA1B are at the object-level rather than the part-level, we apply mask IoU based NMS and use area as NMS score to eliminate part-level object annotations. For GRIT~\cite{kosmos2} data, we extract 5,000,000 samples for scale-up training to enhance the richness of object descriptions.

\begin{table}
\renewcommand\arraystretch{1.0}
\newcommand{\band}{\rowcolor{gray!10}}
    \small 
    \centering
    \resizebox{1.0\columnwidth}{!}{
    \begin{tabular}{lcccccc}
    \toprule
    &\multicolumn{2}{c}{Sizes} & \multicolumn{4}{c}{Annotations}\\
    \cmidrule(lr){2-3} \cmidrule(lr){4-7}
    dataset & images & objects & semantic & box & mask & track id \\
    \midrule
    \band \textbf{Detection Data} &  &  &  &  &  &   \\
    Objects365~\cite{objects365} & 1817287 & 26563198 & category & \checkmark & - & - \\
    OpenImages~\cite{OpenImages} & 1743042 & 14610091 & category & \checkmark & - & - \\
    LVIS~\cite{lvis} & 100170 & 1270141 & category & \checkmark & \checkmark & - \\
    COCO~\cite{coco} & 118287 & 860001 & category & \checkmark & \checkmark & - \\
    BDD~\cite{bdd100k} & 69863 & 1274792 & category & \checkmark & \checkmark & - \\
    \band \textbf{Grounding Data} &  &  &  &  &  &   \\
    RefCOCO~\cite{RefCOCOandplus} & 16994 & 42404 & description & \checkmark & \checkmark & - \\
    RefCOCOg~\cite{RefCOCOg-umd} & 21899 & 42226 & description & \checkmark & \checkmark & - \\
    RefCOCO+~\cite{RefCOCOandplus} & 16992 & 42278 & description & \checkmark & \checkmark & - \\
    VisualGenome~\cite{visualgenome} & 77396 & 3596689 & description & \checkmark & - & - \\
    GRIT~\cite{kosmos2} & 5117307 & 9090607 & description & \checkmark & - & - \\
    \band \textbf{OpenWorld Data} &  &  &  &  &  &   \\
    UVO~\cite{UVO} & 16923 & 157624 & - & \checkmark & \checkmark & - \\
    SA1B~\cite{SAM} & 2147712 & 99427126 & - & \checkmark & \checkmark & - \\
    \band \textbf{Video Data} &  &  &  &  &  &   \\
    YTVIS19~\cite{MaskTrackRCNN} & 61845 & 97110 & category & \checkmark & \checkmark & \checkmark \\
    YTVIS21~\cite{ytvis21dataset} & 90160 & 175384 & category & \checkmark & \checkmark & \checkmark \\
    OVIS~\cite{ovis} & 42149 & 206092 & category & \checkmark & \checkmark & \checkmark \\
    UVO-dense~\cite{UVO} & 45270 & 657990 & - & \checkmark & \checkmark & \checkmark \\
    VOS~\cite{youtubevos} & 94588 & 156310 & - & \checkmark & \checkmark & \checkmark \\
    RefVOS~\cite{urvos} & 93857 & 159961 & description & \checkmark & \checkmark & \checkmark \\
    
    \bottomrule
    \end{tabular}}
    \caption{The tasks \methodNAME learns to complete and the datasets used in training.}
    \label{table:dataset_list}
\end{table}

\begin{table*}[ht]

\begin{center}
\resizebox{\linewidth}{!}{
\begin{tabular}{l@{\hskip9pt}| 
c@{\hskip9pt}c@{\hskip9pt}c@{\hskip9pt} 
c@{\hskip9pt}c@{\hskip9pt}c@{\hskip9pt}
c@{\hskip9pt}c@{\hskip9pt}c@{\hskip9pt}c@{\hskip9pt}
c@{\hskip9pt}l@{\hskip9pt}c@{\hskip9pt}c@{\hskip9pt}cl@{\hskip9pt}l@{\hskip9pt}}
\toprule

Datasets  & \small{OpenImages} &
\small{Objects365} & 
\small{LVIS} &
\small{VisualGenome} &
\small{COCO} &
\small{RefCOCO-mixed} &
\small{SA1B} &
\small{UVO-frame} &
\small{BDD} &
\small{YTVIS19} &
\small{YTVIS21} &
\small{OVIS} &
\small{Ref-YTBVOS} & 
\\
\midrule

Ratio 
& 1.5
& 1.5
& 1.5
& 2
&1.5
& 2.5
& 2.5
& 0.2
& 0.15
& 0.3
& 0.3
& 0.3
& 0.3
  
\\

\bottomrule
\end{tabular}
}
\caption{
The data sampling ratios during the joint-training of stage 2.
RefCOCO-mixed refers to the mixed dataset of RefCOCO~\cite{RefCOCOandplus}, RefCOCO+~\cite{RefCOCOandplus}, RefCOCOg~\cite{RefCOCOg-umd}, and the last four video datasets are treated as independent image data for training.
}
\label{table:data_ratio}
\end{center}
\end{table*}

\textbf{Model and Training Details.}
Following the image backbone, text encoder, and visual prompter, we incorporate a 6-layer deformable transformer encoder and a 9-layer decoder to serve as our Object Decoder following MaskDINO~\cite{maskdino}. We adopt 300 object queries, query denoising, and hybrid matching to accelerate convergence and improve performance. During the pretraining phase of stage 1, we sample data from Objects365 and OpenImages in a 1:1 ratio, with the batch size of 128 for 500,000 training iterations. Moving to stage 2, we train \methodNAME for 500,000 iterations on all image-level data jointly according to the ratios outlined in Table~\ref{table:data_ratio}. For the scale-up training, we set the sampling ratios for SA1B and GRIT to 5.0 in Table~\ref{table:data_ratio}, and train for an extra 500,000 iterations.
We used AdamW~\cite{AdamW} optimizer with base learning rate of $1\times 10^{-4}$,  and weight decay of 0.05, learning rate is decayed at the 400,000 iterations by a factor of 0.1. Learning rates of the image backbone and text encoder are multiplied by a factor of 0.1.  
For the ResNet-50 backbone and Swin backbone, we use scale augmentation~\cite{wu2019detectron2}, resizing the input images such that the shortest side is at least 480 and at most 800 pixels while the longest at most 1333. For EVA02-L backbone, we use the large-scale jittering (LSJ)~\cite{LSJaug} augmentation with a random scale sampled from range 0.1 to 2.0 followed by a fixed size crop to 1536×1536.

\section{Transfer to Video Tasks}
\label{sec:supp_videotask}

To substantiate the effectiveness of \methodNAME across diverse object-level video tasks, we present the performance on VOS and RVOS tasks in Table~\ref{table:vos} and Table~\ref{table:rvos} respectively.

\begin{table}[!ht]
\centering
\resizebox{1.0\linewidth}{!}{
\setlength{\tabcolsep}{2pt}
\begin{tabular}{llcccccccc}
\toprule 
\multirow{2}{*}{\ \ \ \ }&\multirow{2}{*}{Method} & \multicolumn{5}{c}{YT-VOS 2018 val~\cite{youtubevos}}  & \multicolumn{3}{c}{MOSE val~\cite{mose}} \\
\cmidrule(lr){3-7} \cmidrule(lr){8-10}
& & \mg & \mjs & \mfs & \mju & \mfu & \mjf & \mj & \mf\\

\midrule

\parbox[t]{2mm}{\multirow{4}{*}{\rotatebox[origin=c]{90}{Memory}}}
&STM~\cite{stm}&79.4&79.7&84.2&72.8&80.9&-&-&-\tabularnewline
&SWEM~\cite{lin2022swem} &82.8 &82.4 &86.9 &77.1 &85.0 
 &50.9&46.8&64.9\tabularnewline
&STCN~\cite{stcn}&83.0&81.9&86.5&77.9&85.7&50.8&46.6 &55.0\tabularnewline
&XMem~\cite{xmem}&86.1&85.1&89.8&80.3&89.2&57.6&53.3 &62.0\tabularnewline
\midrule

\parbox{2mm}{\multirow{7}{*}{\rotatebox[origin=c]{90}{Non-Memory}}}
&SiamMask~\cite{SiamMask}&52.8&60.2&58.2&45.1&47.7&-&-&-\tabularnewline
&Siam R-CNN~\cite{SiamRCNN}&73.2&73.5&-&66.2&-&-&-&-\tabularnewline
&TVOS~\cite{TVOS}&67.8&67.1&69.4&63.0&71.6&-&-&-\tabularnewline
&FRTM~\cite{FRTM}&72.1&72.3&76.2&65.9&74.1&-&-&-\tabularnewline
&{UNINEXT-R50}~\cite{UNINEXT}& 77.0 & 76.8 & 81.0 &  {70.8} &  {79.4} &-&-&- \tabularnewline
&{UNINEXT-L}~\cite{UNINEXT}&  {78.1} &  {79.1} &  {83.5} &  {71.0} & 78.9 &-&-&-\tabularnewline
&{UNINEXT-H}~\cite{UNINEXT}&  {78.6} &  {79.9} &  {84.9} & 70.6 &  {79.2} &-&-&- \tabularnewline
&\textbf{\methodNAME-Lite } &80.4 &80.2 &85.5 &74.3 &81.4  &56.1&51.8&60.4\tabularnewline
\bottomrule
\end{tabular}}
\caption{
Performance comparison of our \methodNAME on video object segmentation tasks.
}
\label{table:vos}
\end{table}

\textbf{VOS.}
Video object segmentation (VOS) aims at segmenting a particular object throughout the entire video clip sequence. We evaluate \methodNAME on semi-supervised VOS~\cite{caelles2017one} that gives the first-frame mask of the target object on YouTube-VOS 2018~\cite{youtubevos} and MOSE~\cite{mose}.
Given the first-frame mask of the target object, we first crop the prompt square area from RGB image and send it to the image backbone to obtain the visual prompt feature of the corresponding area, and send it to the early fusion module before the Transformer encoder. Then we sample fine-grained visual embeddings from the pixel embedding map $M_p$ inside the given mask area and make them interacted with object queries through self-attention module in the Transformer decoder layer.
We conduct fine-tuning of \methodNAME-Lite jointly on YouTube-VOS~\cite{youtubevos}, YTVIS2019~\cite{MaskTrackRCNN}, YTVIS2021~\cite{ytvis21dataset}, OVIS~\cite{ovis}, and UVO-video~\cite{UVO} for 40,000 iterations. The evaluation is performed on YouTube-VOS and MOSE, as shown in the Table~\ref{table:vos}. It is noteworthy that semi-supervised VOS is almost dominated by space-time memory networks~\cite{stm,lin2022swem,stcn,xmem} which construct a memory bank for each object in the video. \methodNAME achieves the best results among all non-memory-based methods on YouTube-VOS and even demonstrating competitive results compared to memory-based methods on the more challenging MOSE dataset.

\textbf{RVOS.}
Referring Video Object Segmentation (R-VOS) aims at finding objects matched with the given language expressions in a given video and segment them.
Ref-YouTube-VOS~\cite{urvos} is a popular R-VOS benchmarks, which are constructed by introducing language expressions for the objects in the original YouTube-VOS~\cite{youtubevos} dataset.  
As same as semi-supervised VOS, region similarity \mj , contour accuracy \mf, and the averaged score \mjf\ are adopted as the metrics. 
Given an object expression and a video, we send the description into the text encoder, select the object query with the highest confidence score and compute its mask. Additionally, we introduce temporal consistency by adding the similarity between the 300 object queries of the current frame and the object query selected in the previous frame to the current confidence score. We directly evaluate the \methodNAME trained from stage 2 on Ref-YouTube-VOS. As shown in Table~\ref{table:rvos}, \methodNAME outperforms all previous R-VOS approaches and unified method.

\begin{table}[!ht]
\centering
\resizebox{1.0\linewidth}{!}{
\begin{tabular}{l  c  c c c }

\toprule
 {Method} &  {Backbone}   & $\mathcal{J}\&\mathcal{F}$ & $\mathcal{J}$ & $\mathcal{F}$  \\
 \midrule
CMSA~\cite{CMSA}  &\multirow{4}{*}{ResNet-50}& 36.4 & 34.8 & 38.1  \\ 
YOFO~\cite{YOFO}& &48.6&47.5&49.7\\
ReferFormer~\cite{referformer} &  & {58.7} & {57.4} & {60.1}  \\
{UNINEXT}~\cite{UNINEXT}& & {61.2} & {59.3}& {63.0}  \\
\midrule
 
PMINet + CFBI ~\cite{PMINet} & \multirow{2}{*}{Ensemble} & 54.2 & 53.0 & 55.5 \\
CITD ~\cite{CITD} &  & 61.4 & 60.0 & 62.7  \\
 \midrule
ReferFormer~\cite{referformer} & \multirow{2}{*}{Video-Swin-B}  & 64.9 & 62.8 & 67.0   \\
SOC~\cite{luo2023soc} & &67.3 &65.3 &69.3\\
 \midrule
{UNINEXT}~\cite{UNINEXT}&ConvNext-L&{66.2}&{64.0}&{68.4} \\
{UNINEXT}~\cite{UNINEXT}&ViT-H&{70.1} & {67.6} & {72.7} \\
\textbf{\methodNAME-Plus} &Swin-L &67.7 &65.6 &69.7 \\
\textbf{\methodNAME-Pro} &EVA02-L  &70.6 &68.2 &72.9  \\
\bottomrule
\end{tabular}
}
\caption{
Performance comparison of our \methodNAME on Ref-YouTube-VOS task.
}
\label{table:rvos}
\end{table}

\begin{table*}[h]
\begin{center}
\resizebox{\linewidth}{!}{
\begin{tabular}{l@{\hskip9pt} 
c@{\hskip9pt}c@{\hskip9pt}|l@{\hskip9pt} 
l@{\hskip9pt}l@{\hskip9pt}l@{\hskip9pt}
l@{\hskip9pt}l@{\hskip9pt}l@{\hskip9pt}l@{\hskip9pt}
l@{\hskip9pt}l@{\hskip9pt}l@{\hskip9pt}l@{\hskip9pt}l@{\hskip9pt}l}
\toprule

Model & Shot & Tune & \small{PascalVOC} &
\small{AerialDrone} & 
\small{Aquarium} &
\small{Rabbits} &
\small{EgoHands} &
\small{Mushrooms} &
\small{Packages} &
\small{Raccoon} &
\small{Shellfish} &
\small{Vehicles} &
\small{Pistols} &
\small{Pothole} &
\small{Thermal} & 
Avg
\\\midrule
  
\midrule
DyHead \scriptsize{COCO} & 1 & Full 
& 31.7\std3.1
& 14.3\std2.4
& 13.1\std2.0
& 63.6\std1.4
& 40.9\std7.0
& 67.0\std3.6
& 34.6\std12.1
& 45.9\std3.8
& 10.8\std5.0
& 34.0\std3.3
& 12.0\std10.4
& 6.1\std1.3
& 40.9\std7.4
& 31.9\std{3.3}

\\
DyHead \scriptsize{COCO} & 3 & Full
& 44.1\std0.7
& 19.2\std3.0
& 22.6\std1.3
& 64.8\std1.7
& 54.4\std2.5
& 78.9\std1.3
& 61.6\std10.3
& 50.0\std2.1
& 20.8\std3.5
& 44.9\std1.9
& 34.4\std11.1
& 20.6\std2.4
& 57.9\std2.3
& 44.2\std{0.3} 
\\
DyHead \scriptsize{COCO} & 5 & Full

& 44.9\std1.5
& 22.2\std3.0
& 31.7\std1.0
& 65.2\std1.5
& 55.6\std3.7
& 78.7\std3.9
& 50.1\std13.7
& 48.7\std4.8
& 22.8\std3.3
& 52.0\std1.2
& 39.8\std6.7
& 20.9\std1.5
& 48.0\std2.8
& 44.7\std{1.7}

\\
DyHead \scriptsize{COCO} & 10 & Full

& 48.4\std1.2
& 27.5\std1.4
& 39.3\std2.7
& 62.1\std5.9
& 61.6\std1.4
& 81.7\std3.4
& 58.8\std9.0
& 52.9\std3.2
& 30.1\std3.2
& 54.1\std3.3
& 44.8\std4.9
& 26.7\std2.4
& 63.4\std2.8
  & 50.1\std{1.6}
  \\
DyHead \scriptsize{COCO} & All & Full
& 60.1
& 27.6
& 53.1
& 76.5
& 79.4
& 86.1
& 69.3
& 55.2
& 44.0
& 61.5
& 70.6
& 56.6
& 81.0
  & 63.2

\\
\midrule
 DyHead \scriptsize{O365} & 1 & Full
& 25.8\std{3.0}
& 16.5\std{1.8}
& 15.9\std{2.7}
& 55.7\std{6.0}
& 44.0\std{3.6}
& 66.9\std{3.9}
& 54.2\std{5.7}
& 50.7\std{7.7}
& 14.1\std{3.6}
& 33.0\std{11.0}
& 11.0\std{6.5}
& 8.2\std{4.1}
& 43.2\std{10.0}
  & 33.8\std{3.5}
\\
 DyHead \scriptsize{O365} & 3 & Full
& 40.4\std{1.0}
& 20.5\std{4.0}
& 26.5\std{1.3}
& 57.9\std{2.0}
& 53.9\std{2.5}
& 76.5\std{2.3}
& 62.6\std{13.3}
& 52.5\std{5.0}
& 22.4\std{1.7}
& 47.4\std{2.0}
& 30.1\std{6.9}
& 19.7\std{1.5}
& 57.0\std{2.3}
  & 43.6\std{1.0}
\\
 DyHead \scriptsize{O365} & 5 & Full
& 43.5\std{1.0}
& 25.3\std{1.8}
& 35.8\std{0.5}
& 63.0\std{1.0}
& 56.2\std{3.9}
& 76.8\std{5.9}
& 62.5\std{8.7}
& 46.6\std{3.1}
& 28.8\std{2.2}
& 51.2\std{2.2}
& 38.7\std{4.1}
& 21.0\std{1.4}
& 53.4\std{5.2}
  & 46.4\std{1.1}
\\
 DyHead \scriptsize{O365} & 10 & Full
& 46.6\std{0.3}
& 29.0\std{2.8}
& 41.7\std{1.0}
& 65.2\std{2.5}
& 62.5\std{0.8}
& 85.4\std{2.2}
& 67.9\std{4.5}
& 47.9\std{2.2}
& 28.6\std{5.0}
& 53.8\std{1.0}
& 39.2\std{4.9}
& 27.9\std{2.3}
& 64.1\std{2.6}
  & 50.8\std{1.3}
\\
 DyHead \scriptsize{O365} & All & Full
& 53.3 
& 28.4 
& 49.5 
& 73.5 
& 77.9 
& 84.0 
& 69.2 
& 56.2 
& 43.6 
& 59.2 
& 68.9 
& 53.7 
& 73.7 
  & 60.8
\\

 \midrule \midrule
 GLIP-T & 1 & Full
 & 54.8\std{2.0}
& 18.4\std{1.0}
& 33.8\std{1.1}
& 70.1\std{2.9}
& 64.2\std{1.8}
& 83.7\std{3.0}
& 70.8\std{2.1}
& 56.2\std{1.8}
& 22.9\std{0.2}
& 56.6\std{0.5}
& 59.9\std{0.4}
& 18.9\std{1.3}
& 54.5\std{2.7}
  & 51.1\std{0.1}
 
 \\

 GLIP-T & 3 & Full
 & 58.1\std{0.5}
& 22.9\std{1.3}
& 40.8\std{0.9}
& 65.7\std{1.6}
& 66.0\std{0.2}
& 84.7\std{0.5}
& 65.7\std{2.8}
& 62.6\std{1.4}
& 27.2\std{2.7}
& 61.9\std{1.8}
& 60.7\std{0.2}
& 27.1\std{1.2}
& 70.4\std{2.5}
  & 54.9\std{0.2}
 
 \\
 GLIP-T & 5 & Full
 & 59.5\std{0.4}
& 23.8\std{0.9}
& 43.6\std{1.4}
& 68.7\std{1.3}
& 66.1\std{0.6}
& 85.4\std{0.4}
& 72.3\std{0.0} 
& 62.1\std{2.0}
& 27.3\std{1.2}
& 61.0\std{1.8}
& 62.7\std{1.6}
& 34.5\std{0.5}
& 66.6\std{2.3}
  & 56.4\std{0.4}
 
 \\
 GLIP-T & 10 & Full
 & 59.1\std{1.3}
& 26.3\std{1.1}
& 46.3\std{1.6}
& 67.3\std{1.5}
& 67.1\std{0.7}
& 87.8\std{0.5}
& 72.3\std{0.0} 
& 57.7\std{1.7}
& 34.6\std{1.7}
& 65.4\std{1.4}
& 61.6\std{1.0}
& 39.3\std{1.0}
& 74.7\std{2.3}
  & 58.4\std{0.2}
 
 \\
 GLIP-T & All & Full
 & 62.3 
& 31.2 
& 52.5 
& 70.8 
& 78.7 
& 88.1 
& 75.6 
& 61.4 
& 51.4 
& 65.3 
& 71.2 
& 58.7 
& 76.7 
  & 64.9
 \\
\midrule
 GLIP-L & 1 & Full
 & 64.8\std{0.6}
& 18.7\std{0.6}
& 39.5\std{1.2}
& 70.0\std{1.5}
& 70.5\std{0.2}
& 69.8\std{18.0}
& 70.6\std{4.0}
& 68.4\std{1.2}
& 71.0\std{1.3}
& 65.4\std{1.1}
& 68.1\std{0.2}
& 28.9\std{2.9}
& 72.9\std{4.7}
  & 59.9\std{1.4}
 
 \\ 
 GLIP-L & 3 & Full
 
 & 65.6\std{0.6}
& 22.3\std{1.1}
& 45.2\std{0.4}
& 72.3\std{1.4}
& 70.4\std{0.4}
& 81.6\std{13.3}
& 71.8\std{0.3}
& 65.3\std{1.6}
& 67.6\std{1.0}
& 66.7\std{0.9}
& 68.1\std{0.3}
& 37.0\std{1.9}
& 73.1\std{3.3}
  & 62.1\std{0.7}
 \\ 
 GLIP-L & 5 & Full
 
 & 66.6\std{0.4}
& 26.4\std{2.5}
& 49.5\std{1.1}
& 70.7\std{0.2}
& 71.9\std{0.2}
& 88.1\std{0.0} 
& 71.1\std{0.6}
& 68.8\std{1.2}
& 68.5\std{1.7}
& 70.0\std{0.9}
& 68.3\std{0.5}
& 39.9\std{1.4}
& 75.2\std{2.7}
  & 64.2\std{0.3}

 \\ 
 GLIP-L & 10 & Full
 & 66.4\std{0.7}
& 32.0\std{1.4}
& 52.3\std{1.1}
& 70.6\std{0.7}
& 72.4\std{0.3}
& 88.1\std{0.0}
& 67.1\std{3.6}
& 64.7\std{3.1}
& 69.4\std{1.4}
& 71.5\std{0.8}
& 68.4\std{0.7}
& 44.3\std{0.6}
& 76.3\std{1.1}
  & 64.9\std{0.7}
 
 \\
 GLIP-L & All & Full
 & 69.6 
& 32.6 
& 56.6 
& 76.4 
& 79.4 
& 88.1 
& 67.1 
& 69.4 
& 65.8 
& 71.6 
& 75.7 
& 60.3 
& 83.1 
  & 68.9
 
 \\
 
 \midrule\midrule
\methodNAME-Lite  & 1 & Full
&61.3\std{0.5}
&19.2\std{3.1}
&27.2\std{3.4}
&70.8\std{3.3}
&52.8\std{15.1}
&70.7\std{7.5}
&49.2\std{22.0}
&58.1\std{5.4}
&28.8\std{11.0}
&57.9\std{10.0}
&57.7\std{0.6}
&22.2\std{7.9}
&57.0\std{4.5}
&48.7\std{0.9}
\\

\methodNAME-Lite  & 3 & Full
&62.6\std{0.1}
&25.5\std{3.8}
&29.1\std{1.5}
&72.9\std{4.1}
&65.8\std{1.7}
&83.0\std{4.4}
&66.8\std{3.4}
&61.7\std{10.4}
&40.0\std{3.0}
&61.2\std{3.5}
&44.9\std{12.9}
&26.7\std{3.5}
&64.5\std{6.8}
&54.2\std{2.3}
\\
\methodNAME-Lite  & 5 & Full
&62.8\std{0.4}
&28.0\std{3.1}
&33.8\std{2.2}
&71.7\std{2.7}
&64.0\std{4.4}
&81.6\std{4.1}
&64.9\std{5.2}
&60.1\std{12.4}
&39.1\std{1.0}
&59.7\std{3.0}
&49.2\std{14.5}
&30.8\std{1.3}
&69.2\std{7.8}
&55.0\std{3.7}
\\

\methodNAME-Lite  & 10 & Full
&62.1\std{0.9}	
&32.0\std{1.6}	
&39.3\std{2.0}	
&71.2\std{1.5}	
&64.4\std{2.7}	
&88.0\std{2.7}	
&64.3\std{9.8}	
&65.5\std{1.5}	
&36.4\std{4.2}	
&62.1\std{3.4}	
&54.8\std{10.9}	
&38.8\std{1.2}	
&70.6\std{4.0}
&57.7\std{0.6}
\\

\methodNAME-Lite  & All & Full
&62.8
&37.9
&52.9
&73.6
&76.5
&88.9	
&69.7
&65.0 
&51.1	
&58.9
&67.4
&57.2
&82.3
&64.9
\\

\midrule

\methodNAME-Plus  & 1 & Full
&68.2\std{2.2}
&20.4\std{0.2}
&43.9\std{4.1}
&75.5\std{1.6}
&68.4\std{2.7}
&50.6\std{29.0}
&47.3\std{0.8}
&70.4\std{4.0}
&64.6\std{0.5}
&67.7\std{1.5}
&62.3\std{1.0}
&30.0\std{9.2}
&71.6\std{7.7}
&57.0\std{0.8}

\\

\methodNAME-Plus  & 3 & Full
&70.6\std{0.9}	
&24.8\std{2.1}	
&47.6\std{0.8}	
&79.5\std{0.7}	
&69.0\std{2.0}	
&83.1\std{5.9}	
&66.2\std{1.3}	
&75.6\std{3.5}	
&65.3\std{1.1}	
&69.0\std{0.8}	
&65.7\std{4.2}	
&38.1\std{3.1}	
&76.3\std{4.6}
&63.9\std{1.2}	

\\

\methodNAME-Plus  & 5 & Full
&69.9\std{0.9}	
&29.6\std{2.9}	
&48.8\std{1.2}	
&75.0\std{1.7}	
&67.7\std{5.1}	
&83.6\std{9.9}	
&68.5\std{3.2}	
&71.6\std{5.9}	
&61.6\std{4.0}	
&67.7\std{0.8}	
&66.8\std{4.5}	
&38.8\std{1.9}	
&78.9\std{1.0}
&63.7\std{1.0}	
\\

\methodNAME-Plus  & 10 & Full
&69.3\std{1.2}	
&32.5\std{1.9}	
&50.8\std{0.9}	
&76.4\std{0.6}	
&70.7\std{0.9}	
&88.2\std{1.2}	
&68.9\std{3.3}	
&68.2\std{3.0}	
&60.0\std{1.9}	
&69.3\std{1.5}	
&62.6\std{10.3}	
&41.7\std{3.1}	
&81.7\std{1.7}
&64.6\std{1.7}
\\

\methodNAME-Plus  & All & Full
&70.4
&34.8
&54.1
&76.4
&74.5
&89.7
&68.6
&67.6
&57.8
&69.2
&71.4
&57.1
&82.9
&67.3
\\

\midrule

\methodNAME-Pro  & 1 & Full
&70.9\std{1.7}	
&24.5\std{2.3}	
&46.7\std{0.4}	
&76.4\std{0.8}	
&68.2\std{3.8}	
&60.4\std{7.8}	
&58.9\std{2.7}	
&68.2\std{4.5}	
&58.5\std{8.8}	
&67.6\std{0.8}	
&69.2\std{0.2}	
&31.8\std{2.6}	
&70.8\std{7.6}
&59.4\std{1.5}
\\

\methodNAME-Pro  & 3 & Full
&72.3\std{0.4}
&28.4\std{0.5}
&49.6\std{2.2}
&76.1\std{1.3}
&69.3\std{3.9}
&79.4\std{9.5}
&67.4\std{3.5}
&74.1\std{4.9}
&63.7\std{2.0}
&68.4\std{0.6}
&68.3\std{2.1}
&42.1\std{5.3}
&76.9\std{1.6}
&64.3\std{1.3}
\\

\methodNAME-Pro  & 5 & Full
&71.4\std{0.9}
&33.4\std{1.5}
&50.6\std{4.3}
&73.8\std{3.9}
&71.9\std{0.3}
&83.6\std{6.8}
&66.6\std{1.8}
&72.5\std{4.3}
&59.1\std{4.8}
&68.7\std{1.4}
&69.7\std{1.3}
&39.5\std{4.8}
&77.4\std{3.2}
&64.5\std{0.9}
\\

\methodNAME-Pro  & 10 & Full
&71.1\std{1.9}
&37.8\std{2.1}
&54.2\std{1.2}
&73.9\std{7.2}
&70.7\std{1.3}
&90.9\std{1.4}
&66.0\std{9.4}
&73.9\std{6.8}
&57.8\std{3.9}
&69.4\std{0.9}
&62.9\std{6.3}
&44.3\std{3.8}
&79.8\std{0.6}
&65.6\std{0.4}
\\

\methodNAME-Pro  & All & Full
&72.6
&36.5
&58.1
&80.5
&74.1
&92.0
&67.0
&76.5
&66.4
&70.5
&66.4
&55.7
&80.6
&69.0
\\
 \bottomrule
\end{tabular}
}
\caption{Per-dataset performance compared with DyHead, GLIP-T, and GLIP-L. For PascalVOC, we report the mAP (IoU=0.50:0.95) using the COCO evaluation script, to be consistent with other 12 datasets. ``Full'' denotes full-model tuning.}
\label{table:odinw_finetune}
\end{center}
\end{table*}

\section{Object Detection in the Wild}
\label{sec:supp_odinw}
To further validate transferability of \methodNAME on diverse real-world detection tasks, we assess its few-shot transfer ability on the ODinW~\cite{ODinW} dataset. We vary the amount of task-specific annotated data from X-shot, providing at least X examples per category, to using all the available data in the training set, following the procedure established by GLIP~\cite{GLIP}. We fine-tune the models on the provided data using the same hyper-parameters across all models in a full-model tuning regime.
For manually designed prompts, we revise the category names for the two datasets (``Cottontail-Rabbit" to ``rabbit" and ``Cow/Chanterelle" to ``Cow/Chanterelle mushroom") to provide language guidance. Models train with a batch size of 4 and a learning rate of $1\times 10^{-4}$, undergoing 200, 300, 400, 600, and 2000 iterations for the 1, 3, 5, 10, and ALL shot splits, respectively. The optimal model is selected based on the validation split for each train/val split. For each few-shot setting, we train the models three times using different random seeds for train/val splits, and the average score and standard deviation on the test split are reported, as shown in the Table~\ref{table:odinw_finetune}.

\section{Interactive Segmentation and Tracking}
\label{sec:supp_interactive}

\begin{figure*}[tb]
\centering
\includegraphics[width=1.0\linewidth]{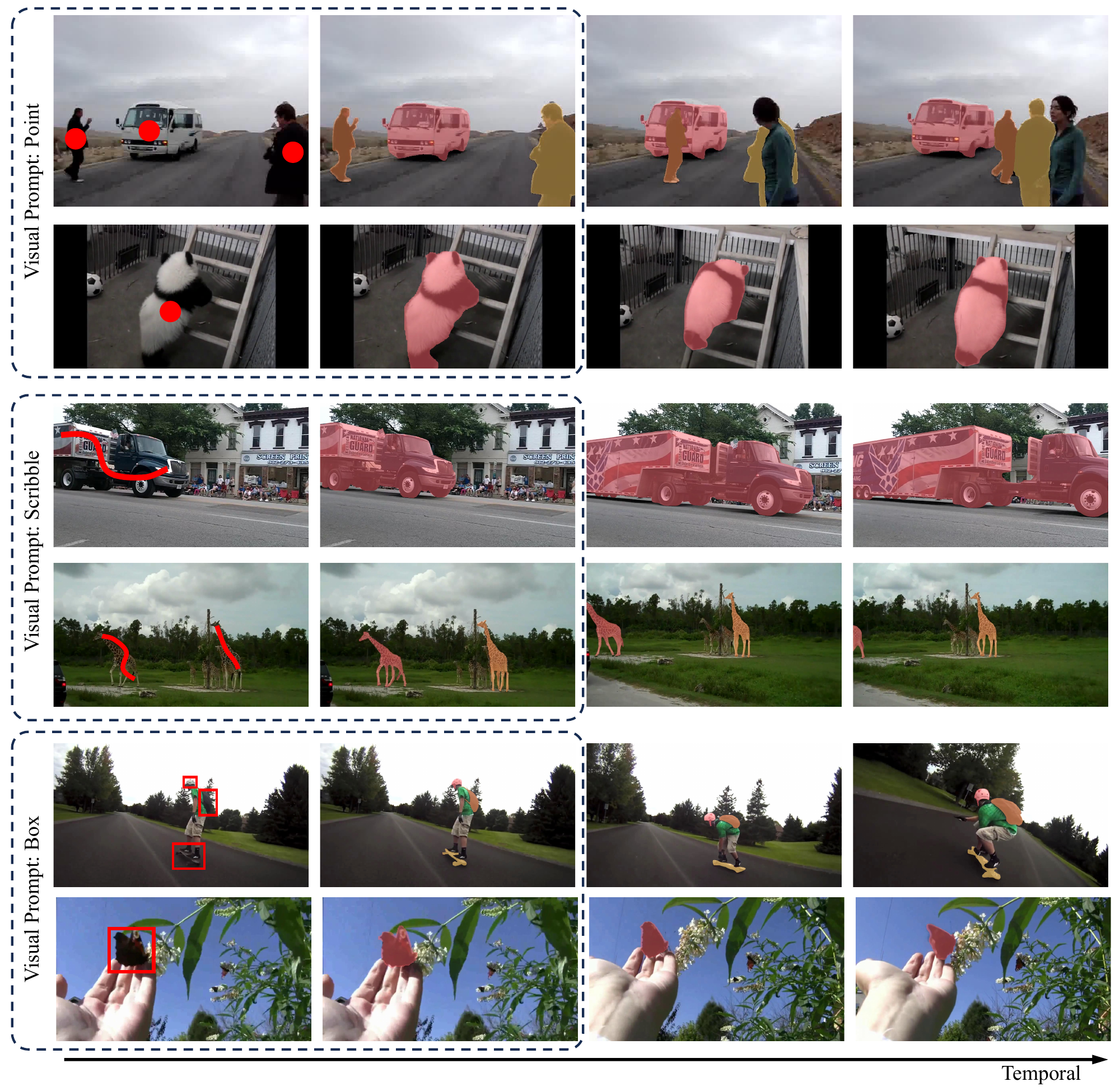}
\caption{
The visualization results of interactive segmentation and tracking. For image-level interactive segmentation, \methodNAME supports sending points, boxes, or scribbles as a visual prompts to the model, enabling direct segmentation of the specified object. In the case of video object segmentation, using the masked feature from the first frame as a prompt referring features allows segmentation of the corresponding object in subsequent frames of the video.
}
\label{fig:visualization}
\end{figure*}

As described in Sec~\ref{sec:supp_videotask}, \methodNAME achieves interactive segmentation and tracking by introducing a visual prompt. Sending points, boxes, or scribbles along with the image to the model enables the segmentation of specified objects. Moreover, by feeding the mask from the previous frame and its corresponding prompt feature into early fusion and self-attention, \methodNAME performs segmentation in the current frame based on the segmentation results from the previous frame. The features of objects in the previous frame serve as referring features at this point. As illustrated in the Figure~\ref{fig:visualization}, we showcase the interactive segmentation results of different prompts on images and videos. Please visit our project homepage to experience more custom interactive image and video segmentation effects through our online demo.

\end{document}